\documentclass[10pt,twocolumn,letterpaper]{article}

\usepackage[pagenumbers]{cvpr} %

\usepackage[accsupp]{axessibility}

\usepackage[dvipsnames]{xcolor}

\usepackage{tabularx}
\usepackage{comment}
\definecolor{cvprblue}{rgb}{0.21,0.49,0.74}
\usepackage[breaklinks,colorlinks,citecolor=cvprblue]{hyperref}
\usepackage{multirow}
\usepackage{array}  %
\usepackage{algorithm}
\usepackage{algpseudocode}

\makeatletter
\@namedef{ver@everyshi.sty}{}
\makeatother
\usepackage{tikz}

\newif\ifdraft
\draftfalse
\drafttrue %
\ifdraft
 \newcommand{\MK}[1]{{\color{magenta}{\bf MK: #1}}}
 
 \newcommand{\JM}[1]{{\color{cyan}{\bf JM: #1}}}
 
 \newcommand{\DN}[1]{{\color{purple}{\bf DN: #1}}}
 \newcommand{\HB}[1]{{\color{red}{\bf HB: #1}}}
 \newcommand{\hp}[1]{{\color{orange}{\bf HP: #1}}}
\else
 \newcommand{\MK}[1]{}
 
 \newcommand{\JM}[1]{}
 
 \newcommand{\DN}[1]{}
 \newcommand{\HB}[1]{}
 \newcommand{\hp}[1]{}
\fi

\newcommand{\Ped}{{\bf Ped2}}
\newcommand{\ST}{{\bf ShanghaiTech}}
\newcommand{\Avenue}{{\bf Avenue}}
\newcommand{\UCF}{{\bf UCFCrime}}
\newcommand{\Ubnormal}{{\bf UBnormal}}

\newcommand{\highlight}[1]{%
  \setlength{\fboxsep}{0pt}\colorbox{gray!30}{$\displaystyle#1$}}

\makeatletter
\renewcommand{\paragraph}{%
  \@startsection{paragraph}{4}%
  {\z@}{2.5ex \@plus 0.75ex \@minus .2ex}{-1em}%
  {\normalfont\normalsize\bfseries}%
}
\makeatother

\def\paperID{14667} %
\def\confName{CVPR}
\def\confYear{2024}

\title{MULDE: Multiscale Log-Density Estimation via Denoising Score Matching for Video Anomaly Detection}

\newcommand{\spacy}{0.75em}
\author{Jakub Micorek \hspace{\spacy} Horst Possegger  \hspace{\spacy}  Dominik Narnhofer \hspace{\spacy} Horst Bischof \hspace{\spacy} Mateusz Kozi\'nski \\
Graz University of Technology, Austria\\
{\tt\small \{jakub.micorek, possegger, dominik.narnhofer, bischof, mateusz.kozinski\}@icg.tugraz.at}
}

\usepackage{fancyhdr}

\newcommand\figvspace{\vspace{-0.0cm}}
\newcommand\figcaptionvspace{\vspace{-0.5cm}}
\newcommand\tabvspace{\vspace{-0.4cm}}
\newcommand\tabcaptionvspace{\vspace{-0.0cm}}

\begin{document}
\maketitle

\begin{abstract}
We propose a novel approach to video anomaly detection: we treat feature vectors extracted from videos as realizations of a random variable with a fixed distribution and model this distribution with a neural network.
This lets us estimate the likelihood of test videos and detect video anomalies by thresholding the likelihood estimates.
We train our video anomaly detector using a modification of denoising score matching, a method that injects training data with noise to facilitate modeling its distribution. 
To eliminate hyperparameter selection, we model the distribution of noisy video features across a range of noise levels and 
introduce a regularizer that tends to align the models for different levels of noise.
At test time, we combine anomaly indications at multiple noise scales with a Gaussian mixture model. 
Running our video anomaly detector induces minimal delays as inference requires merely extracting the features and forward-propagating them through a 
shallow
neural network and a Gaussian mixture model.
Our experiments on five popular video anomaly detection benchmarks demonstrate state-of-the-art performance, both in the object-centric and in the frame-centric setup.
\end{abstract}

\section{Introduction}
The goal of video anomaly detection (VAD) is to detect events that deviate from normal patterns in videos.
VAD has numerous potential applications in healthcare, safety, and traffic monitoring. 
It can be used to detect events like human falling down,
workplace, or traffic accidents, and holds the promise of dramatically reducing the time needed to respond to emergencies that can result from them.
The main challenge of anomaly detection stems from the fact that, unlike classes of actions in video action recognition, anomalies do not form a coherent group of patterns and typically cannot be anticipated in advance.
In consequence, in many applications anomalous training data is not available, necessitating the so-called one-class classification approach, in which the system is trained exclusively on normal data. 

\begin{figure*}
  \center
    \includegraphics[width=\linewidth]{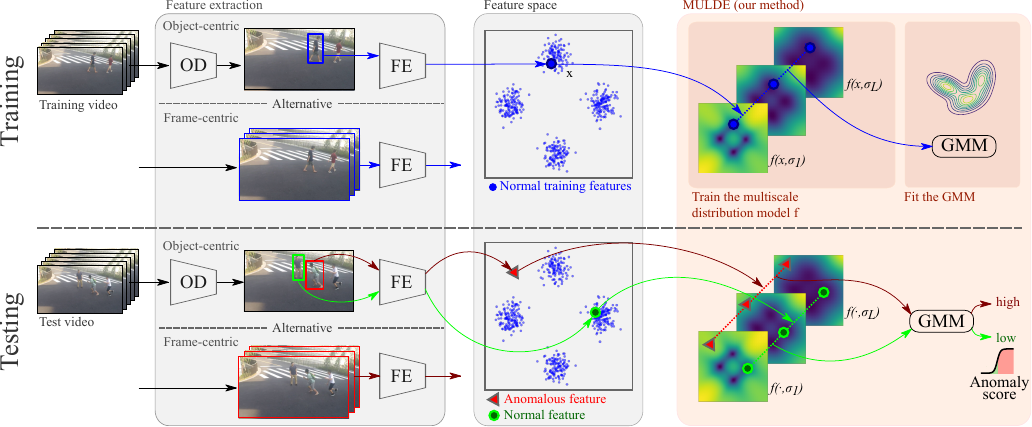}
\figcaptionvspace
  \caption{
MULDE approximates the negative log-density of noisy, normal video features at multiple levels of noise $\sigma$ with a neural network $f(\cdot,\sigma)$.
The log-likelihoods estimated at multiple noise levels are combined into a single anomaly score with a Gaussian mixture model (GMM).
MULDE can be trained to detect video anomalies in an object-centric or frame-centric manner.
In the object-centric approach, an object detector (OD) is used to detect objects which are then fed to the feature extractor (FE).
In the frame-centric approach, the feature extractor is applied to short sequences of entire frames.
}
\label{fig:flowchart}
\figvspace
\end{figure*}

Traditional approaches to one-class video anomaly detection rely on training a deep network in auxiliary self-supervised tasks, like auto-encoding the frame sequence~\cite{georgescu2021anomaly, georgescu2021background, hasan2016learning, ionescu2019object,gong2019memorizing, park2020learning}, predicting future frames \cite{liu2018future, nguyen2019anomaly}, inpainting spatio-temporal volumes~\cite{georgescu2021anomaly}, and solving jigsaw puzzles~\cite{wang2022video,BARBALAU2023103656}.
The underlying assumption is that given a video sufficiently different from those of the training set, \ie one containing an anomaly, the network should fail to complete the self-supervised task.
However, the connection between data normality or abnormality and the performance of the network remains unclear. 
Deep networks can generalize beyond their training set, and there is no guarantee that anomalies make them fail to complete their task.

Our motivation is to lay more solid foundations for video anomaly detection. 
To that end, we treat feature vectors extracted from videos as realizations of a random variable with a fixed distribution, 
and seek to approximate its probability density function with a neural network.
Such approximation would enable a principled and intuitive approach to detecting anomalies:
since anomalous data is characterized by a low likelihood under the statistical model of normal data, it could be detected by thresholding the approximate density function. 

Training a neural network to directly approximate $p(\mathbf{x})$, the probability density function of the training data is very challenging.
However, \citet{vincent2011connection} showed that injecting the data with zero-centered, iid Gaussian noise makes it easier to model the distribution of the noisy data $q(\tilde{\mathbf{x}})$. 
For sufficiently low levels of noise, $q$ preserves the shape of $p$, which makes it a suitable basis for our anomaly indicator. 
Vincent's contribution consisted in proposing denoising score matching,
a method to train a neural network to approximate $-\nabla_{\tilde{\mathbf{x}}} \log q(\tilde{\mathbf{x}})$, the negative log-gradient of the density function of noisy data,
which became a core algorithm of a recent class of generative models~\cite{song2019generative}.
We modify this method to train a neural anomaly indicator that approximates, up to a constant, the log-density, $-\log q(\tilde{\mathbf{x}})$, 
well suited to indicating anomalies thanks to its one-to-one relation to $q(\tilde{\mathbf{x}})$.
Our approach is illustrated in Figure~\ref{fig:flowchart}.

In its basic form, introduced above, our method requires choosing the standard deviation $\sigma$ of the noise injected into the data, also called the noise scale.
This choice represents a compromise between making $q$ closer to $p$ for small values of $\sigma$ and extending the support of $q$ to cover more possible anomalies at larger noise levels.
To avoid this unwelcome compromise,
we do not settle on a single $\sigma$, but approximate the log-density for a range of noise scales $\sigma\in\{\sigma_1,\ldots\sigma_L\}$
with a neural network $f(\cdot,\sigma)$, 
and introduce a regularization term that tends to align the approximations at different scales.
At test time, we compute anomaly indicators for a range of noise scales and combine them into a single anomaly score with a Gaussian mixture model, fitted to normal data.  
Our experiments show that \textbf{MULDE}, the regularized \textbf{MU}ltiscale \textbf{L}og-\textbf{DE}nsity approximation, is a very effective video anomaly detector.

To summarize, the main contribution of this paper is a novel approach to detecting anomalies from video features with a neural approximation of their log-density function.
In technical terms, we propose a modification of multiscale denoising score matching for training anomaly indicators and a new method to regularize this training.
Our anomaly detector is simple, mathematically sound, and fast at test time, as inference requires merely extracting the features and forward-propagating them through a neural network and a Gaussian mixture model.
Moreover, it is agnostic of the feature vector it consumes on input.
Our experiments on the Ped2~\cite{ped2_mahadevan2010anomaly}, Avenue~\cite{avenue_lu2013abnormal}, ShanghaiTech~\cite{shanghaitech_luo2017revisit}, UCFCrime~\cite{ucfcrime_sultani_2018_real_world_vad}, and UBnormal~\cite{ubnormal_Acsintoae_CVPR_2022} data sets demonstrate state-of-the-art performance in anomaly detection
both in the object-centric setup, where features are extracted from bounding boxes of detected objects, and in the frame-centric setup, with features computed for entire frames.

\section{Related Work}
\label{sec:related}

VAD was studied in multiple settings:
as a one-class classification problem, where no anomalous data is available for training~\cite{reiss2022attribute,chaudhry2009histograms,avenue_lu2013abnormal,colque2016histograms,wang2019gods,Yan_2023_ICCV,Flaborea_2023_ICCV,Hirschorn_2023_ICCV,chang2020clustering, hasan2016learning, shanghaitech_luo2017revisit, nguyen2019anomaly, park2020learning, yu2020cloze,yang2022dynamic,feng2021convolutional, zaheer2022generative},
as an unsupervised learning task, where anomalies are present in the training set, but it is not known which training videos contain them~\cite{zaheer2022generative},
and as a supervised, or weakly supervised problem, where training labels indicate anomalous video frames, or videos containing anomalies, respectively~\cite{ucfcrime_sultani_2018_real_world_vad, zaheer2022generative, ubnormal_Acsintoae_CVPR_2022}. 
We address the first of these settings -- we assume the training set is limited to normal videos.

Existing VAD methods can be categorized as frame-centric when they operate on features computed from entire frames or their sequences~\cite{ucfcrime_sultani_2018_real_world_vad,wang2019gods,ubnormal_Acsintoae_CVPR_2022,Yan_2023_ICCV,zaheer2022generative}, and object-centric, if they estimate the abnormality of each bounding box in every frame~\cite{BARBALAU2023103656,georgescu2021anomaly,georgescu2021background,ionescu2019object,reiss2022attribute,wang2022video,Flaborea_2023_ICCV}, typically using a pre-trained feature extractor.
The frame-centric design is more suited for global events, like fires, or smoke, while the object-centric one is oriented at anomalies associated with people or objects, like human falls, or vehicle accidents.
In Sec.~\ref{sec:experiments}, we show that our method can establish state-of-the-art performance with features of either type.

The predominant approach to VAD is to train a deep network to auto-encode normal videos and use the reconstruction error as anomaly indicator~\cite{chang2020clustering, hasan2016learning, shanghaitech_luo2017revisit, nguyen2019anomaly, park2020learning, yu2020cloze,yang2022dynamic,feng2021convolutional}.
The idea of using the error of a model pre-trained on normal data to detect anomalies was extended from auto-encoding to multiple other tasks, including predicting future frames, or the optical flow~\cite{liu2018future,lee2019bman,nguyen2019anomaly,yu2020cloze,feng2021convolutional}, 
inpainting spatio-temporal volumes~\cite{georgescu2021anomaly}, and solving jigsaw puzzles~\cite{wang2022video,BARBALAU2023103656}.
This overarching approach is predicated on the assumption that the error is higher for anomalous frames than for normal frames.
However, there is no certainty that this assumption holds: it is not well understood under what conditions a neural network fails to perform its task and there is no guarantee that all anomalies make it fail.
By contrast, no heuristic assumptions underlie the functioning of MULDE. 

The idea of detecting video anomalies by modeling the distribution of normal video features recurs in the literature, but, to date, effective modeling techniques remain elusive. 
Some methods, like Gaussian mixture models~\cite{reiss2022attribute},
one-class Support Vector Machines~\cite{chaudhry2009histograms,avenue_lu2013abnormal,colque2016histograms},
or multilinear classifiers~\cite{wang2019gods},
may lack the expressive power needed to reflect the complex and high-dimensional distribution of video features.
Adversarially trained models~\cite{lee2018stan,georgescu2021background} offer high expressive power,
but cannot guarantee to fully capture the distribution, because
parts of the feature space may remain unexplored by the generator-discriminator pair during training.
Diffusion models capture data distribution, but their use in VAD consists in generating samples of normal frames~\cite{Yan_2023_ICCV}, or human poses~\cite{Flaborea_2023_ICCV}, and comparing observed frames or poses to generated ones. 
This requires multiple diffusion steps and reduces the anomaly measure to a distance between the observation and the sample.
Normalizing flows, recently used for detecting anomalies in human pose features~\cite{Hirschorn_2023_ICCV}, are free from these drawbacks as they explicitly approximate the likelihood of the training data.
However, their performance decreases in the presence of complex correlations between features~\cite{kirichenko2020why}.
In contrast to these methods, MULDE combines all key ingredients of a VAD approach: high expressive power, the capacity to fully capture the distribution, and to accommodate arbitrary features. 

\begin{figure}
  \centering
  \newcommand{\myfont}{\small}
  \newcommand{\lefttext}{0.16}
  \begin{tikzpicture}
    \node[anchor=south west, inner sep=0] at (0,0) {\includegraphics[width=\linewidth]{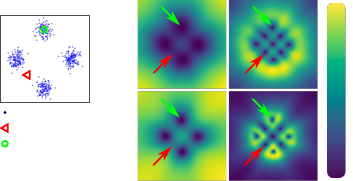}};
    
    \node[anchor=west, align=left, font=\myfont, text=black] at (\lefttext, 1.6) {Training features};
    
    \node[anchor=west, align=left, font=\myfont, text=black] at (\lefttext, 1.25) {Anomalous feature};
    
    \node[anchor=west, align=left, font=\myfont, text=black] at (\lefttext,0.87) {Normal feature};

    \node[anchor=center, align=left, font=\myfont, text=black] at (4.25, 4.5) {$- \log q(x)$};
    
    \node[anchor=center, align=left, font=\myfont, text=black] at (6.45, 4.5) {$\lVert -\nabla_x \log q(x) \rVert$};
    
    \node[anchor=center, align=left, font=\myfont, text=black, rotate=90] at (3,1.1) {Low noise $\sigma$};
    
    \node[anchor=center, align=left, font=\myfont, text=black, rotate=90] at (3,3.2) {Higher noise $\sigma$};

    \node[anchor=center, align=left, font=\myfont, text=white, rotate=90] at (7.96,0.5) {low};
    
    \node[anchor=center, align=left, font=\myfont, text=black, rotate=90] at (7.96,3.7) {high};

  \end{tikzpicture}
  \figcaptionvspace
  \caption{The log-density function is well suited for indicating anomalies, but its gradient is not. (\emph{Left:}) A sample from a mixture of 4 Gaussians. (\emph{Right:}) Learned negative log-density approximation (left column) and the norm of its gradient (right column).
The negative log-density is a good anomaly indicator, taking low values for \textcolor{green!65!black}{normal data} and higher values for \textcolor{red}{anomalous data}.
By contrast, the log-gradient norm is low not only at the modes of the distribution, but also at its minima between the modes, making it impossible to distinguish some anomalies from normal data.
  }
  \label{fig:motivation}
  \figvspace
\end{figure}
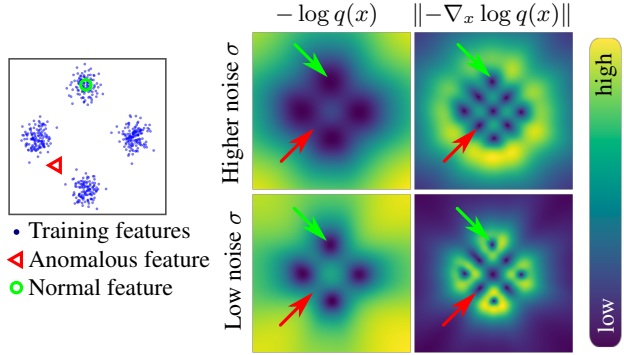

The log-density approximation $f$, employed by MULDE to model the distribution of normal data,
is often called the energy, in reference to the energy-based models~\cite{lecun2006tutorial}, which represent probability distributions in the Boltzmann form
$q(x)=\frac{1}{Z} e^{-f(x)}$, 
but restrict the model to the energy function $f$, since computing the normalization constant $Z$ is typically infeasible.
MULDE can therefore be seen as an energy-based model.
For training $f$, MULDE relies on a modification of denoising score matching~\cite{vincent2011connection}, a method to train a neural network to approximate the energy gradient.
Score matching models the distribution of the training data injected with iid Gaussian noise, and Song and Ermon~\cite{song2019generative} extended it to multiple noise levels.
Mahmood \etal~\cite{mahmood2020multiscale} used the norm of this multi-level energy gradient approximation to detect anomalies in images.
However, the gradient indicates all stationary points of the log-density function, which may appear both at the modes of the distribution, where normal data is concentrated, and in low-density regions, where anomalous data may reside. 
As shown in Fig.~\ref{fig:motivation}, some minima and maxima of the distribution remain indistinguishable even across a range of noise scales. 
By contrast, MULDE approximates the log-density function which, unlike its gradient, is a good anomaly indicator.

\section{Method}
\label{sec:method}
\vspace{-1mm}
We perform anomaly detection in the space of semantic features extracted from videos.
This lets us focus on detecting semantic anomalies, for example, unusual actions involving objects observed also under normal conditions,
as opposed to anomalies in the space of raw input, like frame sequences that do not resemble real videos.
We delegate feature extraction to off-the-shelf models
and focus on the effective detection of anomalous events that they encode. 
\vspace{-1mm}
\paragraph{Motivation}
Intuitively, anomalous video features should not be observed under `normal' conditions.
More formally, an anomalous feature is characterized by 
a small likelihood under the assumed statistical model of anomaly-free video features.
This suggests an approach to detecting anomalies by approximating the probability density function $p$ and simply declaring test features with sufficiently low probability anomalous.
We adopt this approach and model the probability density with a neural network.

\vspace{-1mm}
\paragraph{Overview}
In practice, it is difficult to train a neural network to directly approximate the probability density of its training data $p$.
However, injecting the data with noise makes this approximation feasible, as we will explain below.
The distribution of the noisy data takes the form
\begin{equation} \label{eq:noisy_distribution}
q (\tilde{\mathbf{x}}) = \int \rho(\tilde{\mathbf{x}} \vert \mathbf{x} ) p(\mathbf{x}) d \mathbf{x} ,
\end{equation}
where $\rho(\tilde{\mathbf{x}} \vert \mathbf{x} )$ denotes the conditional distribution of a noisy sample $\tilde{\mathbf{x}}$ given a noise-free sample $\mathbf{x}$, which we take to be an iid Gaussian centered at $\mathbf{x}$.
The main idea behind our anomaly detector is that $q$ preserves the shape of $p$ but, in contrast to it, yields itself to a neural approximation.
Specifically, we approximate, up to a constant, the negative log-density function $-\log q(\tilde{\mathbf{x}})$, which is an excellent anomaly indicator due to its bijective relation to $q (\tilde{\mathbf{x}})$.
Low values of $-\log q(\tilde{\mathbf{x}})$ correspond to high probability density and are characteristic of normal data.
Its high values indicate areas of low probability density where anomalous data may reside. 
Fig.~\ref{fig:motivation} illustrates this on a synthetic example.
The technique we use to approximate the log-density with a neural network is a modification of score matching~\cite{vincent2011connection}, a method to approximate the log-density gradient.

We introduce score matching in Sec.~\ref{sec:method:score_matching} and our training method in Sec.~\ref{sec:method:training}.
In Sec.~\ref{sec:method:multiscale} and~\ref{sec:method:regularization}, we extend this method to approximating the negative log-density at different scales of injected noise, and introduce a regularization term intended to facilitate combining the multi-scale approximations.
We discuss the choice of video features in Sec.~\ref{sec:method:features}.
A diagram of our approach is presented in Fig.~\ref{fig:flowchart}.

\subsection{Background: Denoising score matching}
\label{sec:method:score_matching}
\citet{vincent2011connection} proposed a method to train a neural network $s$, parameterized with a vector $\theta$, to approximate the gradient of the negative log-density function of data perturbed with iid Gaussian noise, in the sense of solving
\begin{equation} \label{eq:noised_score_matching}
\min_\theta \mathbb{E}_{\mathbf{\tilde{x}}\sim q(\mathbf{\tilde{x}})} \left\| s_{\theta}(\mathbf{\tilde{x}}) + \nabla_{\mathbf{\tilde{x}}} \log q(\mathbf{\tilde{x}}) \right\|_2^2 .
\end{equation}
This gradient approximation forms the foundation of a family of generative models~\cite{song2019generative} which initialize a sample with noise and use the log-gradient approximation to drive the sample close to the mode of the distribution.
We will show that it also enables effective detection of video anomalies.

Directly evaluating the objective~\eqref{eq:noised_score_matching} is impossible because $q(\mathbf{\tilde{x}})$ is not known analytically, but~\citet{vincent2011connection} showed that it is equivalent, up to a constant, to
\begin{equation} \label{eq:score_matching_gaussian}
\min_\theta
\mathbb{E}_{\substack{\mathbf{x}            \sim p(\mathbf{x}) \\
                      \mathbf{\tilde{x}}    \sim \mathcal{N}(\mathbf{\tilde{x}} | \mathbf{x},\sigma \mathbf{I})}}
\left\| s_{\theta}(\mathbf{\tilde{x}}) - 
  \frac{\mathbf{\tilde{x}} - \mathbf{x}}{\sigma^{2}} 
  \right\|_2^2 ,
\end{equation}
which can be evaluated effectively.
This gives rise to a stochastic algorithm for training $s$ that iterates: composing a batch of noise-free training data $\mathbf{x}$, perturbing it with Gaussian noise to obtain a batch of noisy data $\tilde{\mathbf{x}}$, and making a gradient step on the expectation in Eq.~\eqref{eq:score_matching_gaussian}, evaluated for the batch.
Since this resembles training $s$ to predict the noise injected to $\mathbf{x}$, it is often called \emph{denoising} score matching.

\subsection{Anomaly detection by denoising score matching}
\label{sec:method:training}
We modify the denoising score matching formulation to train a neural network $f_\theta$, where $\theta$ denotes the vector of parameters, to approximate $-\log q(\tilde{\mathbf{x}})$, as opposed to its gradient.
To that end, we change the objective~\eqref{eq:score_matching_gaussian}, to train the \emph{gradient} of $f$, instead of the network itself, which yields 
\begin{equation}
\label{eq:training_gradient}
\min_\theta
\mathbb{E}_{\hspace{-3mm}\substack{\mathbf{x}            \sim p(\mathbf{x}) \\
                      \mathbf{\tilde{x}}    \sim \mathcal{N}(\tilde{\mathbf{x}} | \mathbf{x},\sigma \mathbf{I})}}
   \left\| 
      \nabla_{\mathbf{\tilde{x}}} f_\theta \left(\mathbf{\tilde{x}}\right)
      -
      \frac{\mathbf{\tilde{x}} - \mathbf{x}}{\sigma^2}
   \right\|_2^2
.
\end{equation}
This makes $\nabla_{\mathbf{\tilde{x}}} f_\theta \left(\mathbf{\tilde{x}}\right)$ approximate $-\nabla_{\tilde{\mathbf{x}}}\log q(\tilde{\mathbf{x}})$ and, by the fundamental theorem of calculus, aligns $f$ with $-\log q(\tilde{\mathbf{x}})$ up to a constant.
Here, $f: \mathbb{R}^d \rightarrow \mathbb{R}$ is a mapping from the space of video features to scalar log-density values, and $\nabla_{\mathbf{\tilde{x}}} f_{\theta}: \mathbb{R}^d \rightarrow \mathbb{R}^d$, like $s$ in the standard score matching formulation~\eqref{eq:score_matching_gaussian}, maps $d$-dimensional video features to $d$-dimensional vectors of log-density gradients.

Notably, our formulation has an advantage over the standard denoising score matching even when the goal is to approximate the gradient of the log-density, as opposed to the log-density itself.
By the Stokes' theorem, gradients form conservative -- that is, curl-free -- vector fields.
Directly approximating the log-density gradient with a neural network, as done by the standard approach, may result in a vector field that is not conservative, in other words, does not represent a gradient of any function. 
By contrast, the gradient of a neural network trained using our formulation is guaranteed to form a conservative vector field. 
On the downside, training $f_{\theta}$ with our loss requires the network to be twice differentiable,
which precludes the use of ReLU, several other nonlinearities, and max pooling.

\subsection{Distribution modeling across noise scales}
\label{sec:method:multiscale}
We recall that training with our loss~\eqref{eq:training_gradient} makes $f_{\theta}$ approximate the negative log-density of the distribution of noisy data $q(\tilde{\mathbf{x}})$, connected to the distribution of noise-free data $p(\mathbf{x})$ through the noise distribution $\rho(\tilde{\mathbf{x}} \vert \mathbf{x} )$. 
$\rho$ is an iid Gaussian centered at $\mathbf{x}$ and with a standard deviation $\sigma$.
The choice of $\sigma$, called the noise scale, represents a compromise between making $q$ closer to $p$ at small noise scales and extending its support to cover more anomalies for larger noise levels.
In theory, the optimal noise scale could be selected by cross-validation on a combination of normal and anomalous data,
but in practice, anomalous validation data is rarely available. 
Therefore, instead of settling for a single noise scale, we approximate the log-density for a range of noise scales 
and combine the estimates at different scales by modeling their joint distribution with a Gaussian mixture.

To implement the multiscale log-density approximation, we take inspiration from~\citet{song2019generative}, who extended the original score matching formulation, presented in Eq.~\eqref{eq:score_matching_gaussian}, to multiple noise scales, and apply a similar extension to our objective~\eqref{eq:training_gradient}.
Instead of approximating $-\log q(\tilde{\mathbf{x}})$ for a fixed $\sigma$, we approximate a family of functions $-\log q_\sigma (\tilde{\mathbf{x}})$, parameterized by $\sigma$, with a neural network $f_{\theta}$, conditioned on $\sigma$.
We found it beneficial to put more emphasis on smaller values of $\sigma$ when training $f_{\theta}$.
Thus, we sample $\sigma$ from the log-uniform distribution on the interval $[\sigma_\text{low},\sigma_\text{high}]$ and minimize
\begin{equation}
\label{eq:training_multiscale}
\min_\theta
\mathbb{E}_{\hspace{-3mm}\substack{\mathbf{x}            \sim p(\mathbf{x}) \\
                                   \mathbf{\tilde{x}}    \sim \mathcal{N}(\tilde{\mathbf{x}} | \mathbf{x},\sigma \mathbf{I}) \\
                                   \sigma\sim\mathit{\mathcal{LU}}(\sigma_\text{low},\sigma_\text{high})}}
\!
   \lambda ( \sigma ) 
   \left\| 
      \nabla_{\mathbf{\tilde{x}}} f_\theta \left(\mathbf{\tilde{x}}, \sigma\right)
      -
      \frac{\mathbf{\tilde{x}} - \mathbf{x}}{\sigma^2}
   \right\|_2^2
,
\end{equation}
where $\lambda(\sigma)$ is a factor that balances the influence of the loss terms at different noise levels. We set $\lambda(\sigma)=\sigma^2$. 

Once the network is trained, we fit a Gaussian mixture model to multi-scale log-density approximation vectors $[f_{\theta}(\mathbf{x},\sigma_i)]_{i=1\ldots L}$ for an evenly spaced sequence of noise levels $\left\{ \sigma_i \right\}^L_{i=1}$, where $\sigma_1=\sigma_\text{low}$ and $\sigma_L=\sigma_\text{high}$.
At test time, our neural network takes a vector of video features and produces a multi-scale vector of log-density approximations, which is then input to the Gaussian mixture model yielding the final anomaly score.

\subsection{Multiscale training regularization}
\label{sec:method:regularization}

The limitation of our method is that $f_{\theta}(\cdot,\sigma)$ can be trained to approximate $-\log q_\sigma(\tilde{\mathbf{x}})$ only up to a constant.
That is, $f_{\theta}(\tilde{\mathbf{x}},\sigma)$ effectively approximates $-\log q_\sigma(\tilde{\mathbf{x}})+C_\sigma$, where $C_\sigma$ is a constant that we do not know.
In our formulation, there is no guarantee that this constant does not change across the range of $\sigma$. 
Since the variation of $C_\sigma$ may make it more difficult to aggregate the estimates at different scales,
we discourage it by using a regularization term $f_{\theta}(\mathbf{x}, \sigma)^2$, that penalizes the log-densities of noise-free examples.
Our full training objective thus becomes
\begin{multline}
\label{eq:our_objective}
\!\!\!\!\!\!
\min_\theta
\mathbb{E}_{\hspace{-3mm}\substack{\mathbf{x}            \sim p(\mathbf{x}) \\
                      \mathbf{\tilde{x}}    \sim \mathcal{N}(\tilde{\mathbf{x}} | \mathbf{x},\sigma \mathbf{I}) \\
                      \sigma\sim\mathit{\mathcal{LU}}(\sigma_\text{low},\sigma_\text{high})}}
\Big[
   \lambda ( \sigma ) 
   \left\| 
      \nabla_{\mathbf{\tilde{x}}} f_\theta \left(\mathbf{\tilde{x}}, \sigma\right) 
      -
      \frac{\mathbf{\tilde{x}} - \mathbf{x}}{\sigma^2}
   \right\|_2^2 \\ 
   + \beta f_{\theta}(\mathbf{x}, \sigma)^2
\Big]
,
\end{multline}
where $\beta$ is a hyperparameter of our method.
Minimizing~\eqref{eq:our_objective} no longer makes $\nabla_\mathbf{x} f_{\theta}$ an unbiased estimate of the log-gradient of the distribution, but as shown in our ablation studies, it improves our results in video anomaly detection.
Algorithm~\ref{alg:training} summarizes training the log-density approximation $f_\theta$.
The Gaussian mixture model is fitted with the standard expectation-maximization algorithm.

\begin{algorithm}
\small
\caption{\small Training MULDE's anomaly indicator. The terms and steps that differ from the standard multi-scale denoising score matching~\cite{song2019generative} are highlighted.}
\begin{algorithmic}[1]

\Require 
    \Statex $f_\theta$ neural log-density model, parameterized by $\theta$
    \Statex $\mathcal{T}$ training set of normal video features
    \Statex $\sigma_\text{low}$ and $\sigma_\text{high}$, limits of the noise scale range 
    \Statex $\beta$ regularization strength
\vspace{0.25em}
\State $\theta \gets$ random initialization 
\While{not converged} 
    \State $\mathcal{X} \gets$ sample a batch from $\mathcal{T}$ 
    \For {$\mathbf{x} \in \mathcal{X}$} \Comment{for each batch element}
        \State $\sigma \gets \text{sample log-uniform}(\sigma_{\text{low}}, \sigma_{\text{high}})$
        \State $\mathbf{\tilde{x}} \gets$ sample $\mathcal{N}(\mathbf{x},\sigma \mathbf{I})$ 
        \State Compute $\highlight{f_\theta} \left(\mathbf{\tilde{x}}, \sigma\right)$ by forward-propagation %
        \State \hspace*{-\fboxsep}\colorbox{gray!30}{Compute $f_\theta \left(\mathbf{      {x}}, \sigma\right)$ by forward-propagation}
        \State \hspace*{-\fboxsep}\colorbox{gray!30}{Compute $\nabla_{\mathbf{\tilde{x}}} f_\theta \left(\mathbf{\tilde{x}}, \sigma\right)$ by backpropagation }
        \State $\mathcal{L}(\mathbf{x}) \gets \sigma^2 \left\| \highlight{\nabla_{\mathbf{\tilde{x}}} f_\theta} \left(\mathbf{\tilde{x}}, \sigma\right)  - \frac{\mathbf{\tilde{x}} - \mathbf{x}}{\sigma^2} \right\|_2^2 + \highlight{\beta f_\theta \left(\mathbf{      {x}}, \sigma\right)^2}$ %
    \EndFor
    \State $\mathcal{L} \gets  \frac{1}{|\mathcal{X}|} \sum_{\mathbf{x}\in\mathcal{X}} \mathcal{L(\mathbf{x})}$ \Comment{aggregate loss over batch}
    \State Compute $\nabla_{\theta} \mathcal{L}$ by backpropagation
    \State Update $\theta$ using $\nabla_{\theta} \mathcal{L}$ with Adam
\EndWhile
\end{algorithmic}
\label{alg:training}
\end{algorithm}

\subsection{Selection of video features}
\label{sec:method:features}
Feature selection is closely tied to the type of target anomalies.
For example, human pose features are well suited for detecting falls, and optical-flow-based ones help detect objects moving with unusual speeds or in unusual directions. 
MULDE is feature agnostic and in Sec.~\ref{sec:experiments} we demonstrate its application with pose, velocity, and deep features, extracted from bounding boxes of object proposals, and ones extracted from entire frames and their sequences.
As reported in 
Sec.~\ref{sec:experiments}, 
feature extraction dominates the running time of our method, which enables selecting the feature extractor to match the desired frame rate.

\section{Experimental Evaluation}
\label{sec:experiments}

\paragraph{Data sets}
We evaluated MULDE on five VAD benchmarks, containing videos captured with static cameras: 
\begin{itemize}%
  \item \Ped~\cite{ped2_mahadevan2010anomaly} includes 16 training and 12 test videos of a campus scene. The training videos show pedestrians and the test videos contain anomalies, like cyclists, skateboarders, and cars in pedestrian areas.
  \item \Avenue~\cite{avenue_lu2013abnormal} consists of 16 training and 21 test videos of a walkway. Anomalies include people running, throwing objects, and walking in the wrong direction.
  \item \ST~\cite{shanghaitech_luo2017revisit} comprises 330 training and 107 test videos of 13 pedestrian traffic scenes differing by the camera viewpoint and lighting conditions. Anomalous events include robbery, jumping, fighting, and cycling.
  \item \Ubnormal~\cite{ubnormal_Acsintoae_CVPR_2022} is composed of 543 synthetic videos of 29 virtual scenes. It contains human-related anomalies, like fighting, running, and jumping, but it also includes car accidents and environmental anomalies, like fog. 
  \item \UCF~\cite{ucfcrime_sultani_2018_real_world_vad} contains 1900 real-world surveillance videos, totaling 128 hours, and including 13 anomaly types, like fighting, robbery, road accidents, and burglary.
\end{itemize}
All our experiments were run in the `one-class classification' setting, that is, we used no anomalous videos for training.
In the experiments on \Ubnormal{} and \UCF{}, which contain anomalous training videos, we discarded these videos and restricted training to normal data.

\paragraph{Performance metrics}
We followed the standard practice and gauged performance on a frame-by-frame basis.
For the object-centric approaches, which yield an anomaly score for each object detected in every frame, we took the highest, \ie most anomalous, score in a frame for the evaluation.
We used the area under the receiver operating characteristic curve (AUC-ROC) as the main performance metric.  

Two methods to aggregate the AUC-ROC over multiple videos can be found in the literature: the \emph{micro} and the \emph{macro} score~\cite{georgescu2021background,ristea2022self,ubnormal_Acsintoae_CVPR_2022,reiss2022attribute,BARBALAU2023103656}. For the \emph{macro} score, the AUC-ROC is computed separately for each video and then averaged across all videos in the test set. The \emph{micro} score computes the AUC-ROC jointly for all frames of all test videos.
In abstract terms, the \emph{macro} score reflects performance attained by adjusting the detection threshold for each video independently, and the \emph{micro} score is more conservative and applies the same threshold to all test videos.
Since we address a VAD use case without an adaptive threshold, we rely on the \emph{micro} score in our evaluation, but report the results in terms of both metrics.
In the supplementary material, we additionally report the tracking- and region-based metrics by~\citet{ramachandra2020street}.

\paragraph{Baselines}
We compared MULDE to the best-performing VAD algorithms, including ones based on the reconstruction error~\cite{ionescu2019object,yu2020cloze,georgescu2021anomaly}, auxiliary tasks~\cite{liu2021hybrid,ristea2022self,wang2022video,georgescu2021anomaly,ubnormal_Acsintoae_CVPR_2022,BARBALAU2023103656,Yan_2023_ICCV}, adversarial training~\cite{georgescu2021background,feng2021convolutional,ubnormal_Acsintoae_CVPR_2022,zaheer2022generative}, normalizing flows~\cite{Hirschorn_2023_ICCV}, and one-class classification with a multilinear classifier~\cite{wang2019gods}.
We reproduced the performance metrics of these methods as reported in the original papers.
We present an even broader comparison, including less recent work, in the supplementary material.

Two baselines are related to our method more closely than the others.
The {AccI-VAD}~\cite{reiss2022attribute} approximates the probability density function of normal video features with a Gaussian mixture model, while we perform this approximation with a neural network.
{MSMA}~\cite{mahmood2020multiscale} uses the norm of the log-density gradient as an anomaly indicator, while our anomaly indicator is based on the log-density itself. Since {MSMA} was developed for image anomaly detection, we re-implemented it to work with video features.

\paragraph{Implementation details}
In our experiments, our density model $f_{\theta}$ parametrized by $\theta$ has two hidden layers with 4096 units followed by GELU nonlinearities. The final layer has an output dimension of one without any nonlinearity.
It is trained using the Adam update rule~\cite{kingma2014adam}, with exponential decay rates $\beta_1=0.5$ and $\beta_2=0.9$, and a batch size of 2048.
We use the learning rates of 5e-4 and 1e-4 in the object- and frame-centric experiments, respectively.

Both during training and at test time, we standardize the video features component-wise using the statistics of the training set. 
During training, we sample the noise scale used for each batch element from the log-uniform distribution on the interval $[\sigma_{\text{low}}, \sigma_{\text{high}}] = [\text{1e-3}, 1.0]$.
For evaluation, we use $L = 16$ evenly spaced noise levels between $[\sigma_{\text{low}}, \sigma_{\text{high}}]$.

\paragraph{Video feature extraction}
MULDE is agnostic of the features used to represent the video content. In our experiments, we reused off-the-shelf video feature extractors.

In the object-centric experiments, we used the feature extraction pipeline proposed by~\citet{reiss2022attribute}. It detects objects in each video frame and extracts deep and velocity features for each detected object. Additionally, pose features are extracted for each detected person. The pose vector contains coordinates of 17 keypoints and is obtained with AlphaPose~\cite{fang2017rmpe}. CLIP image encoder~\cite{radford2021learning} is used to extract deep features, which take the form of 512-dimensional vectors. The velocity features are produced by FlowNet 2.0~\cite{ilg2017flownet} and binned into histograms of oriented flows.

In the frame-centric experiments, we extracted features using Hiera-L~\cite{hiera_ryali2023hiera}, a masked autoencoder pre-trained on images and fine-tuned on Kinetics400~\cite{kay2017kinetics}, a large-scale video action recognition data set. Hiera-L takes sequences of 16 frames as input and produces feature 1152-dimensional feature vectors.

\paragraph{Performance in object-centric VAD}

\begin{table}
\centering
\footnotesize
\setlength{\tabcolsep}{5pt}
\begin{tabular}{l *{2}{p{.5cm}}  *{2}{p{.5cm}} *{2}{p{.6cm}}}
\toprule
\multirow{2}{*}{Method} & 
\multicolumn{2}{c}{Ped2} & 
\multicolumn{2}{c}{Avenue} & 
\multicolumn{2}{c}{ShanghaiTech} \\
\cmidrule(r){2-3} \cmidrule(r){4-5} \cmidrule(r){6-7}
& Micro & Macro & Micro & Macro & Micro & Macro \\
\midrule
                                        CAE-SVM \cite{ionescu2019object} &             94.3 &             97.8 &             87.4 &             90.4 &               78.7 &               84.9 \\
                                                  VEC \cite{yu2020cloze} &             97.3 &                - &             90.2 &                - &               74.8 &                  - \\
                                       SSMTL \cite{georgescu2021anomaly} &             97.5 & \underline{99.8} &             91.5 &             91.9 &               82.4 &               89.3 \\
                                             HF$^2$ \cite{liu2021hybrid} &             99.3 &                - &             91.1 &             93.5 &               76.2 &                  - \\
                                   BA-AED \cite{georgescu2021background} &             98.7 &             99.7 &             92.3 &             90.4 &               82.7 &               89.3 \\
             \cite{georgescu2021background}+SSPCAB \cite{ristea2022self} &                - &                - &             92.9 &             91.9 &               83.6 &               89.5 \\
                                      Jigsaw-Puzzle \cite{wang2022video} &             99.0 &    \textbf{99.9} &             92.2 &             93.0 &               84.3 &               89.8 \\
\cite{georgescu2021anomaly}+UbNormal \cite{ubnormal_Acsintoae_CVPR_2022} &                - &                - &             93.0 &             93.2 &               83.7 &               90.5 \\
                                      AccI-VAD \cite{reiss2022attribute} &             99.1 &    \textbf{99.9} &             93.3 &    \textbf{96.2} &               85.9 &               89.6 \\
                                       SSMTL++ \cite{BARBALAU2023103656} &                - &                - & \underline{93.7} &             92.5 &               83.8 &               90.5 \\
                                      MSMA* \cite{mahmood2020multiscale} & \underline{99.5} &    \textbf{99.9} &             90.2 &             92.5 &               84.1 &               90.2 \\
                                      STG-NF \cite{Hirschorn_2023_ICCV} &                - &                - &                - &                - &               85.9 &                  - \\
\hline
												MULDE$_{\beta=0}$ (ours) &    \textbf{99.7} &    \textbf{99.9} &             93.1 & \underline{96.1} &   \underline{86.4} &   \underline{91.0} \\
                                                            MULDE (ours) &    \textbf{99.7} &    \textbf{99.9} &    \textbf{94.3} & \underline{96.1} &      \textbf{86.7} &      \textbf{91.5} \\
\bottomrule
\end{tabular}
\tabcaptionvspace
\caption{Object-centric results. Frame-level AUC-ROC (\%) comparison (best marked \textbf{bold}, second best \underline{underlined}).  *implemented by us.}
\label{tab:results_object_centric}
\tabvspace
\end{table}

We present the results in object-centric VAD in Table~\ref{tab:results_object_centric}.
MULDE outperforms all the baselines in terms of the more conservative \emph{micro} score.
Interestingly, {AccI-VAD} which, like MULDE, relies on modeling the probability density of normal data, ranks third on all three data sets. 
The high accuracy of both methods speaks in favor of the density modeling approach, while the edge MULDE holds over {AccI-VAD} attests to the superiority of MULDE's neural density model over the combination of Gaussian mixture models and the k-th nearest neighbor technique employed by {AccI-VAD}.
{MSMA}, which uses an approximation of the log-density gradient as anomaly indicator, is outperformed by our log-density-based method by a fair margin.

\paragraph{Performance in frame-centric VAD}
\begin{table}
\centering
\footnotesize
\setlength{\tabcolsep}{5pt}
\begin{tabular}{l *{2}{p{.5cm}}  *{2}{p{.5cm}} *{2}{p{.6cm}}}

\toprule
\multirow{2}{*}{Method}  & \multicolumn{2}{c}{ShanghaiTech}  &   \multicolumn{2}{c}{UCF-Crime}  &   \multicolumn{2}{c}{UBnormal}  \\
\cmidrule(r){2-3} \cmidrule(r){4-5} \cmidrule(r){6-7}
 & Micro & Macro & Micro & Macro & Micro & Macro \\
\midrule
             CT-D2GAN \cite{feng2021convolutional} &               77.7 &                  - &                - &                - &                - &                - \\
                       CAC~\cite{wang2020cluster} &   \underline{79.3} &                  - &                - &                - &                - &                - \\
                  Scene-Aware~\cite{sun2020scene} &               74.7 &                  - &             72.7 &                - &                - &                - \\
                         GODS~\cite{wang2019gods} &                  - &                  - &             70.5 &                - &                - &                - \\
                  GCL~\cite{zaheer2022generative} &                  - &                  - &             74.2 &                - &                - &                - \\
     UBnormal~\cite{ubnormal_Acsintoae_CVPR_2022} &                  - &                  - &                - &                - &             68.5 &             80.3 \\
                        FPDM~\cite{Yan_2023_ICCV} &               78.6 &                  - &             74.7 &                - &             62.7 &                - \\
AccI-VAD$_{\text{GMM}}$*~\cite{reiss2022attribute} &               76.2 &               82.9 &             60.3 &             84.5 &             66.8 &             83.2 \\
AccI-VAD$_{\text{kNN}}$*~\cite{reiss2022attribute} &               71.9 &               83.1 &             53.0 &             82.7 &             65.2 &             82.5 \\
               MSMA*~\cite{mahmood2020multiscale} &               76.7 &               84.2 &             64.5 &             83.4 &             70.3 &             85.1 \\
\hline
                          MULDE$_{\beta=0}$(ours) &               78.4 &      \textbf{86.0} & \underline{75.9} & \underline{84.8} & \underline{71.3} &    \textbf{86.0} \\
                                      MULDE(ours) &      \textbf{81.3} &   \underline{85.9} &    \textbf{78.5} &    \textbf{84.9} &    \textbf{72.8} & \underline{85.5} \\
\bottomrule                                

\end{tabular}
\tabcaptionvspace
\caption{Frame-centric results. Frame-level AUC-ROC (\%) comparison (best marked \textbf{bold}, second best \underline{underlined}). *implemented by us.}
\label{tab:results_frame_centric}
\tabvspace
\end{table}

As can be seen in Table~\ref{tab:results_frame_centric}, in frame-centric VAD, MULDE outperforms other methods on all three data sets. 
Notably, on the \UCF{} data set, by far the largest publicly available VAD benchmark, we advance the state of the art in terms of the \emph{micro} score by $3.8$ percent points.
We improve the state of the art by more than 2pp on \ST{} and \Ubnormal{}.

Moreover, while recent object-centric methods dominate frame-centric methods on data sets enabling both types of evaluation, MULDE narrows this performance gap, attaining the \emph{micro} score of 81.3\% in frame-centric VAD on \ST{} and 86.7\% in object-centric VAD on the same data set.
On \Ubnormal{}, our frame-centric approach even outperforms the object-centric state-of-the-art method~\cite{Hirschorn_2023_ICCV} by 1.1 percent points of the \emph{micro} score.

\paragraph{Ablation studies}
To validate the design of MULDE, we run ablation studies and evaluated the contribution to performance from our regularization term, the Gaussian mixture model, and the architecture of our anomaly indicator.

\begin{table}
\centering
\footnotesize
\setlength{\tabcolsep}{4pt}
\begin{tabular}{l *{2}{p{.7cm}}  *{2}{p{.7cm}} *{2}{p{.7cm}}}
\toprule
\multirow{2}{*}{$\beta$}  & \multicolumn{2}{c}{ShanghaiTech}  &   \multicolumn{2}{c}{UCF-Crime}  &   \multicolumn{2}{c}{UBnormal}  \\
\cmidrule(r){2-3} \cmidrule(r){4-5} \cmidrule(r){6-7}
 & Micro & Macro & Micro & Macro & Micro & Macro \\
\midrule
$0.0$  &              78.4 &     \textbf{86.0} &              75.9 &              84.8 &              71.3 &     \textbf{86.0} \\
\hline
$0.01$ &              80.7 &              84.9 &              76.6 &  \underline{84.9} &     \textbf{72.9} &              85.2 \\
$0.1$  &  \underline{81.3} &  \underline{85.9} &     \textbf{78.5} &  \underline{84.9} &  \underline{72.8} &  \underline{85.5} \\
$1.0$  &     \textbf{81.4} &              84.5 &  \underline{77.2} &     \textbf{85.5} &              72.5 &              84.7 \\
\bottomrule
\end{tabular}
\tabcaptionvspace
\caption{
Frame-centric performance of MULDE trained with different values of the regularization parameter $\beta$. Frame-level AUC-ROC (\%) comparison (best marked \textbf{bold}, second best \underline{underlined}).
\label{tab:ablation:beta}
}
\tabvspace
\end{table}

\begin{figure}
\centering
\begin{tikzpicture}
\node (image1) at (0,0) {\includegraphics[width=\linewidth]{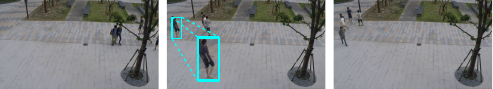}};

\node (image2) at (0,-3.) {\includegraphics[width=\linewidth]{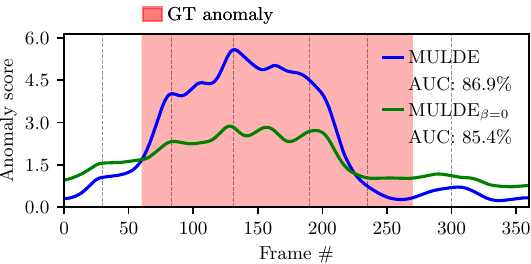}};

\node (image3) at (0,-6) {\includegraphics[width=\linewidth]{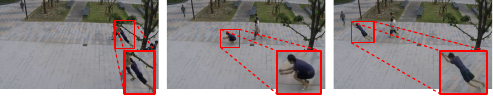}};

\newcommand{\topImagesBottom}{-0.7}
\newcommand{\midPlotTop}{-1.6}
\newcommand{\bottomImagesTop}{-5.3}
\newcommand{\midPlotBottom}{-4.0}

\newcommand{\centerLeft}{-2.8}
\newcommand{\centerCenter}{-0}
\newcommand{\centerRight}{2.8}

\draw[->] (\centerLeft, \topImagesBottom) -- (-2.55, \midPlotTop);
\draw[->, cyan] (\centerCenter, \topImagesBottom) -- (1.6, \midPlotTop);
\draw[->] (\centerRight, \topImagesBottom) -- (2.9, \midPlotTop);

\draw[->, red] (\centerLeft, \bottomImagesTop) -- (-1.5, \midPlotBottom);
\draw[->, red] (\centerCenter, \bottomImagesTop) -- (-0.5, \midPlotBottom);
\draw[->, red] (\centerRight, \bottomImagesTop) -- (0.7, \midPlotBottom);

\end{tikzpicture}
\figcaptionvspace
\vspace{-0.15cm}
\caption{Anomaly detection with MULDE in a test video of the \ST{} data set (video 13 in scene 4). 
Pedestrians walking in frames 30 and 300 represent normal behavior. A person jumping across the scene is annotated as \textcolor{red}{anomalous}.
The anomaly indication produced by MULDE is aligned with the ground truth (GT) at its beginning but terminates earlier than the GT annotation.
However, careful examination of the video reveals that \textcolor{cyan}{normal behavior} (walking, cyan bounding box in the top row) is re-instantiated before the end of the annotation, as indicated by MULDE.
A regularized model produces a stronger anomaly indication (plotted in blue) than one without regularization (green plot). 
}
\label{fig:qualitative:04_0013}
\figvspace
\end{figure}
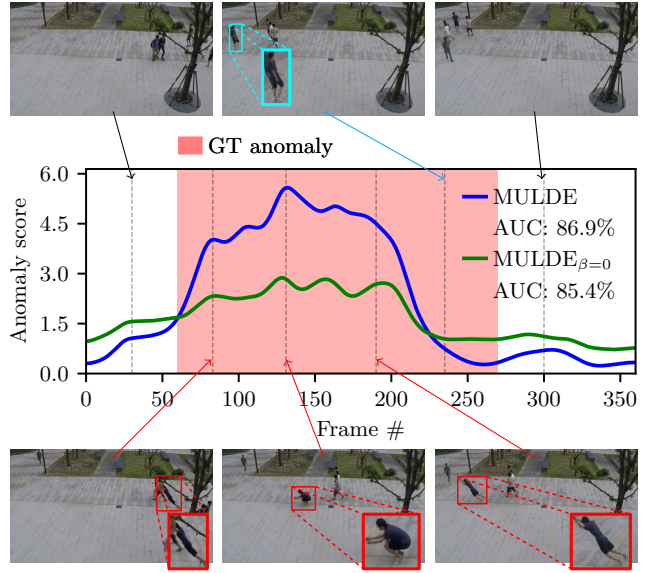

As shown in Table~\ref{tab:ablation:beta}, regularization improves the results of frame-centric VAD in terms of the \emph{micro} score, although different values of the regularization factor $\beta$ are optimal for different data sets. We take $\beta=0.1$ as a compromise among performance on the three data sets.
A qualitative example of anomaly detection by MULDE with and without regularization, shown in Fig.~\ref{fig:qualitative:04_0013}, demonstrates that a regularized model produces stronger anomaly indications.

\begin{figure}
    \centering
    \includegraphics[width=\columnwidth]{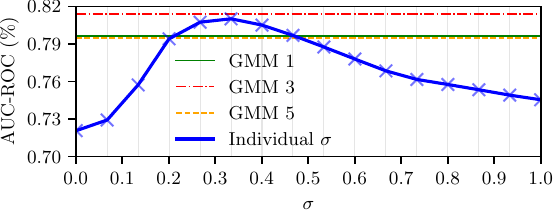}
    \figcaptionvspace
    \vspace{-0.16cm}
    \caption{Performance of MULDE in frame-centric VAD on the \ST{} data set without the Gaussian mixture model, with our anomaly indicator computed at individual noise scales (blue plot), and with the mixture model with 1, 3, and 5 components.
    \label{fig:sigma_sweep_gmm}
    }
\figvspace
\end{figure}

As explained in Section~\ref{sec:method:multiscale}, MULDE combines anomaly indications computed for a range of noise scales with a Gaussian mixture model fitted to normal data. This lets us avoid the need to select the noise scale to be used at test time.
Fig.~\ref{fig:sigma_sweep_gmm} demonstrates the impact of the mixture model on performance. It shows the \emph{micro} score attained by our anomaly indicator $f_{\theta}(\cdot,\sigma)$ at individual noise scales, without the mixture model. 
The best performance is attained for intermediate levels of noise, which suggests a tradeoff between representing the noise-free distribution more faithfully at lower noise scales and better covering the space away from training examples at higher noise scales. Performance attained with the mixture model slightly exceeds the one attained with the optimal noise scale, validating the GMM as a method to avoid tuning the noise level.

\begin{table}
\centering
\footnotesize
\setlength{\tabcolsep}{6pt}
\begin{tabular}{r *{2}{p{.45cm}}  *{2}{p{.45cm}} *{2}{p{.45cm}} *{2}{p{.5cm}}}
\toprule

\multirow{2}{*}{Units} & \multicolumn{2}{c}{1 Layer} & \multicolumn{2}{c}{2 Layers} & \multicolumn{2}{c}{3 Layers} & \multicolumn{2}{c}{4 Layers}  \\
\cmidrule(lr){2-3} \cmidrule(lr){4-5} \cmidrule(lr){6-7} \cmidrule(lr){8-9}
& Micro & Macro & Micro & Macro & Micro & Macro & Micro & Macro \\
\midrule
  1024 &     77.3 &     84.2 &          79.9 &          85.4 & \underline{81.2} & \underline{85.8} &     79.9 &     84.2 \\
  2048 &     77.6 &     84.6 &          80.4 &          84.6 &             79.2 &             83.5 &     79.4 &     84.0 \\
  4096 &     78.4 &     84.6 & \textbf{81.3} & \textbf{85.9} &             79.5 &             84.4 &     80.0 &     84.2 \\
  8192 &     79.2 &     84.3 &          79.4 &          84.8 &             80.4 &             84.4 &     79.3 &     84.7 \\
\bottomrule
\end{tabular}
\caption{
Performance of MULDE with architectural variations in frame-centric VAD on \ST{}. Units indicate the number of neurons within a hidden layer. Columns correspond to the number of hidden layers. Frame-level AUC-ROC (\%) comparison (best marked \textbf{bold}, second best \underline{underlined}).
\label{tab:ablation:nn}
}
\tabvspace
\end{table}

To find the optimal architecture of our anomaly indicator $f_{\theta}$, we performed an ablation study over the number of hidden layers and the number of neurons in each hidden layer. The results, presented in Table~\ref{tab:ablation:nn}, suggest the optimal depth of two hidden layers and the width of 4096 neurons, which we used in our experiments.

\paragraph{Running time}
Once a feature has been extracted from the video, MULDE requires only a forward-pass through the shallow network $f_{\theta}$ and an evaluation of a Gaussian mixture model. This represents a very small computational overhead over feature extraction. In the frame-centric approach with 512-dimensional feature vectors, MULDE takes less than one millisecond to process a single feature.
For comparison, the Hiera-L feature extractor~\cite{hiera_ryali2023hiera}, used in the frame-centric experiments, requires 130 milliseconds to compute one feature vector from a sequence of 16 frames.
MULDE's running time is therefore dominated by the time needed for feature extraction, which means that the feature extractor can be selected to match the target framerate on a given architecture. For example, in our setup, extracting a single feature with the smaller Hiera-base model takes 33 milliseconds, which enables video anomaly detection at 25 FPS, even on our PC with an NVIDIA RTX 2080Ti GPU.

\section{Discussion}

\paragraph{Limitations}
Our method has two main limitations.
First, our video anomaly detector requires features of anomalous and normal videos to be distinguishable.
Intuitively, this condition should be satisfied when the video feature extractor is selected adequately, but there is no theoretical guarantee that this is indeed the case.
Second, even though our method eliminates the need to select the noise scale used for training, the range of employed noise scales should be sufficiently large to cover anomalies that would be seen at test time.
Currently, there is no automatic way to select the upper limit of the noise range. 
We circumvent this limitation by standardizing the features component-wise and using a fixed, wide range of noise scales. 

\paragraph{Future work}

We plan to extend this work by exploring architectures of the video anomaly indicator.
In particular, reformulating the anomaly indicator in terms of the reconstruction error~\cite{georgescu2021anomaly, georgescu2021background, hasan2016learning, ionescu2019object,gong2019memorizing, park2020learning} represents a distinct opportunity to integrate previous methods into our approach.

Our method uses Gaussian noise during training, but other forms of noise may represent anomalies more effectively.
For example, emulating random motion patterns in the video might be useful for detecting objects moving along unusual trajectories or with abnormal velocities. 

Finally, we plan to extend MULDE from the one-class-classification scenario to the (weakly-) supervised one.

\paragraph{Conclusion}
We presented a novel approach to video anomaly detection by modeling the distribution of non-anomalous video features with a neural network.
Our method is fast and attains state-of-the-art performance, both in the frame-centric and object-centric VAD.
Most importantly, it is firmly grounded in statistical modeling.
We defined abnormality in terms of the probability density under the distribution of normal data
and designed MULDE to approximate the log-density function.
This lets us discuss our method in statistical terms.  
For example, when our anomaly indicator fails to classify a video as anomalous, we know that its log-density approximation is not good enough.
We can then plan actions to rectify this, for example, by changing the neural network architecture, acquiring more training data, or modifying the range of noise.
By contrast, it might not be obvious what actions should be taken when previous methods fail, for example, by showing a small reconstruction error for an anomalous item.

\paragraph{Acknowledgements}
This work was partially funded by the Austrian Research Promotion Agency (FFG) under the project High-Scene (884306) and by the Austrian Science Fund (FWF) Lise Meitner grant (M3374).

{
    \small
    \bibliographystyle{ieeenat_fullname}
    \bibliography{refs}
}

\clearpage
\newcommand{\beginsupplement}{%
\setcounter{table}{0}
\renewcommand{\thetable}{S\arabic{table}}%
\setcounter{figure}{0}
\renewcommand{\thefigure}{S\arabic{figure}}%
}
\beginsupplement

\setcounter{page}{1}
\maketitlesupplementary

This supplementary material extends the results presented in the main manuscript with additional visualizations (Section~\ref{sec:vis}) and detailed experiment results (Section~\ref{sec:ablations}).

\section{Visualizations} \label{sec:vis}
\paragraph{Intuitive example of multiscale log-density estimation}
\begin{figure*}
  \centering
  \begin{minipage}[b]{1.\textwidth}
  \centering
  \begin{subfigure}[b]{0.49\textwidth}
    \centering
    \includegraphics[width=0.75\linewidth]{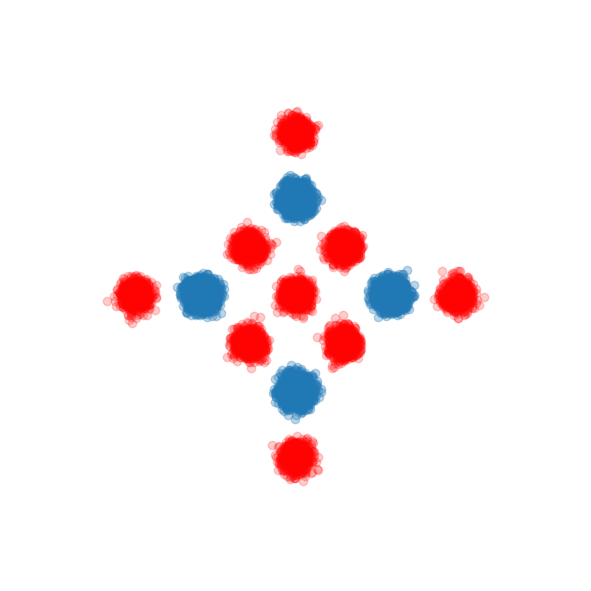}
    \subcaption{
    Example dataset: \textcolor{blue}{Normal features} and \textcolor{red}{anomalous features}.}
    \label{fig:dataset}
  \end{subfigure}
  \hfill
  \begin{subfigure}[b]{0.49\textwidth}
    \centering
    \includegraphics[width=0.85\linewidth]{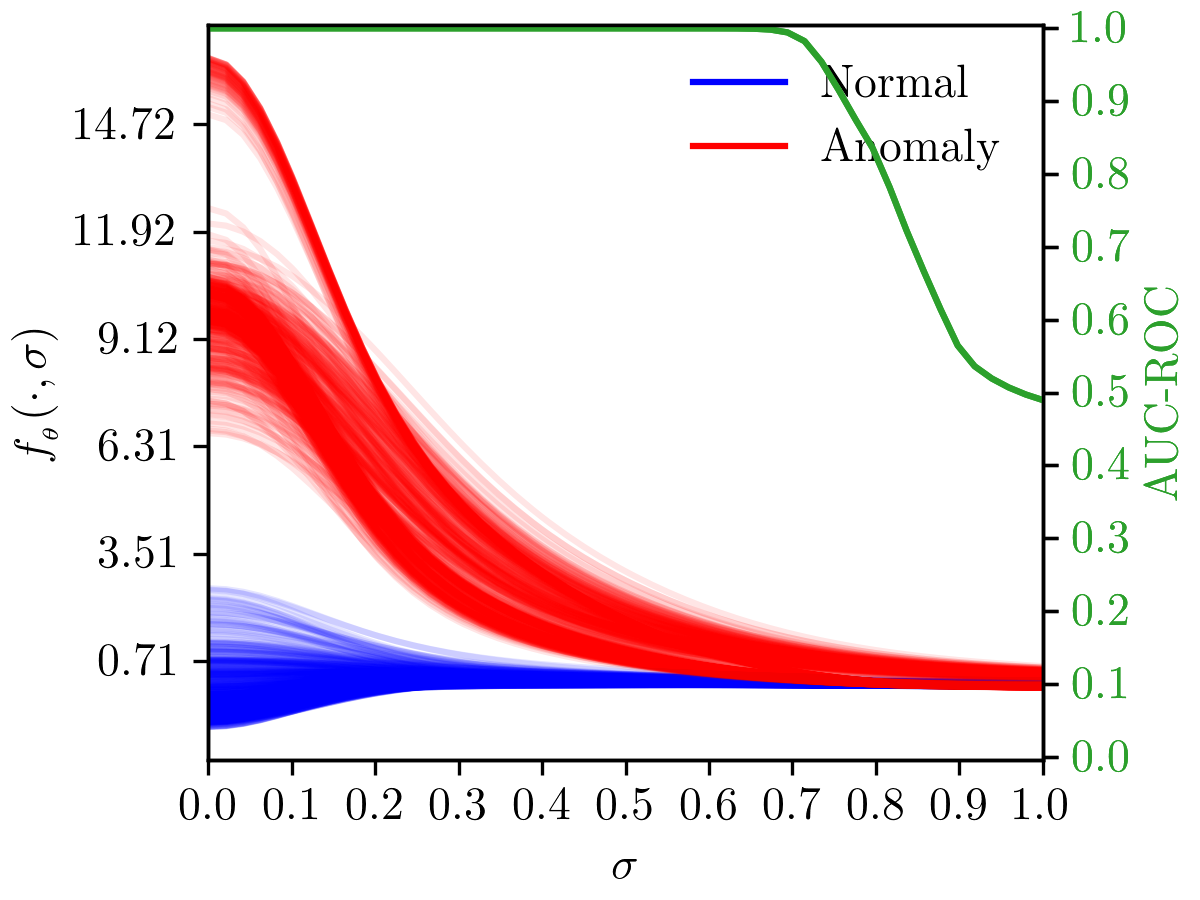}
    \subcaption{AUC-ROC across multiple noise-scales $\sigma$ based on the negative log probability $f_{\theta}$. The \textcolor{blue}{normal} and \textcolor{red}{anomalous} samples are well separable. 
    }
    \label{fig:noise_level_fthetacdotsigma}
  \end{subfigure}
\end{minipage}
    \vspace{0.1cm}

  \begin{minipage}[b]{1.\textwidth}
  \begin{subfigure}[b]{0.24\textwidth}
    \includegraphics[width=\linewidth]{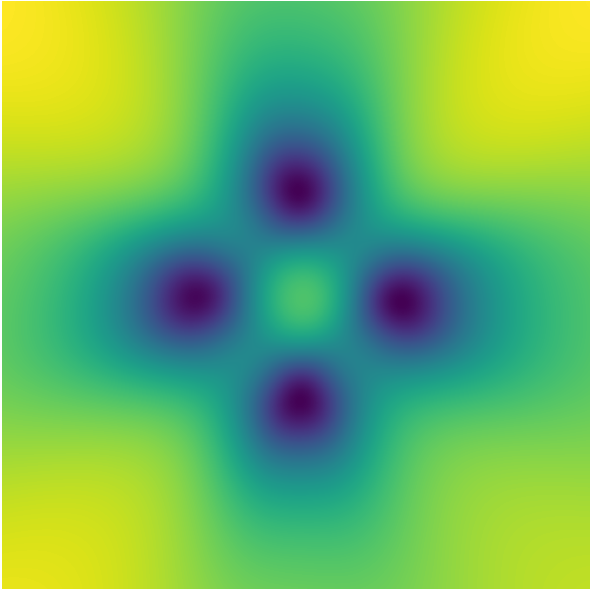}
    \caption*{$f_{\theta}(\cdot, \sigma = 0.001)$}
  \end{subfigure}
  \begin{subfigure}[b]{0.24\textwidth}
    \includegraphics[width=\linewidth]{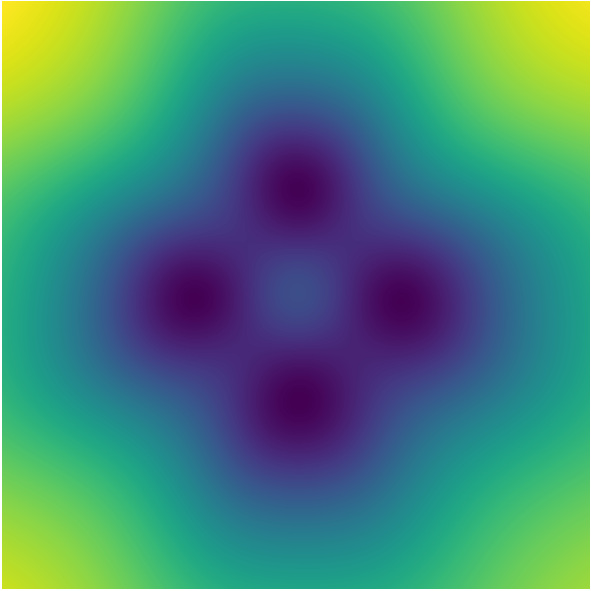}
    \caption*{$f_{\theta}(\cdot, \sigma = 0.1)$}
  \end{subfigure}
  \begin{subfigure}[b]{0.24\textwidth}
    \includegraphics[width=\linewidth]{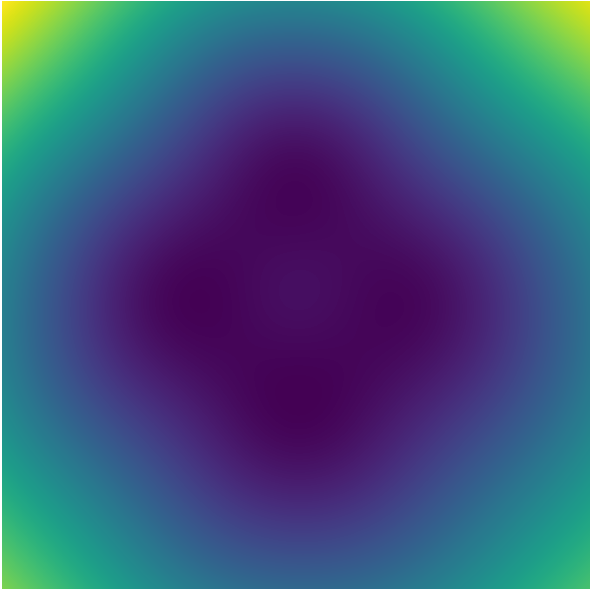}
    \caption*{$f_{\theta}(\cdot, \sigma = 0.5)$}
  \end{subfigure}
  \begin{subfigure}[b]{0.24\textwidth}
    \includegraphics[width=\linewidth]{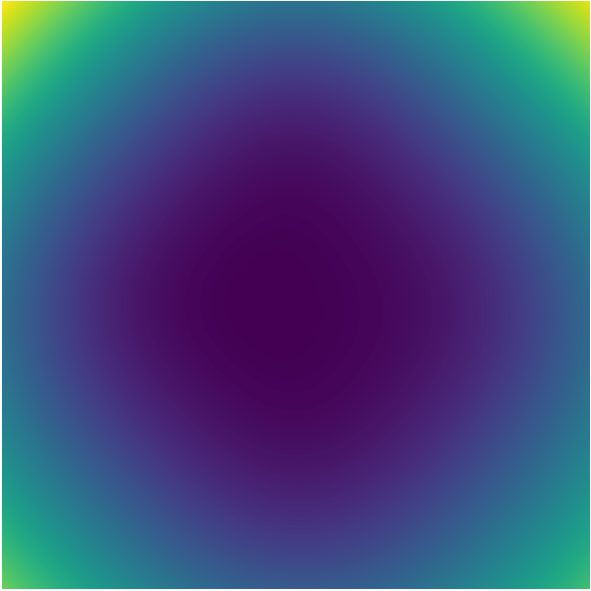}
    \caption*{$f_{\theta}(\cdot, \sigma = 1.0)$}
  \end{subfigure}
  \vspace{0.2cm}
  \subcaption{The log-density of \textcolor{blue}{normal training features} is estimated with $f_{\theta}$ across multiple $\sigma$. MULDE leverages $f_{\theta}$ as a strong anomaly indicator. \label{fig:toyexample_logdensity}
    }
  \end{minipage}  
  
  \caption{The intuition behind the use of log-density estimation for anomaly detection. (a) \textcolor{blue}{Normal features}, sampled from a mixture of four Gaussians, are shown in blue, while \textcolor{red}{anomalous features} are shown in red. (b) compares the values of $f_\theta$ for features from the \textcolor{blue}{normal} and \textcolor{red}{anomalous} samples. Each graph shows the log-density at multiple noise scales for a single sample.
  Our anomaly indicator $f_{\theta}$ is well suited to separate anomalies from normal data. (c) shows the log-density approximations across noise scales.}
  \label{fig:sup:toy}
\end{figure*}

To provide more intuition about our neural log-density approximation, in Figure~\ref{fig:sup:toy} we present a toy example that extends Figure~\ref{fig:motivation} from the main manuscript.
In Figure~\ref{fig:noise_level_fthetacdotsigma}, we plot our log-density approximation as a trajectory across a range of noise scales $\sigma$, for each \textcolor{blue}{normal} and \textcolor{red}{anomalous} sample.
Our log-density estimation
separates normal and anomalous data well across a wide range of noise scales $\sigma$.
We show the log-density estimation in Figure~\ref{fig:toyexample_logdensity}.

\begin{figure*}
\centering
\includegraphics[width=1.\linewidth]{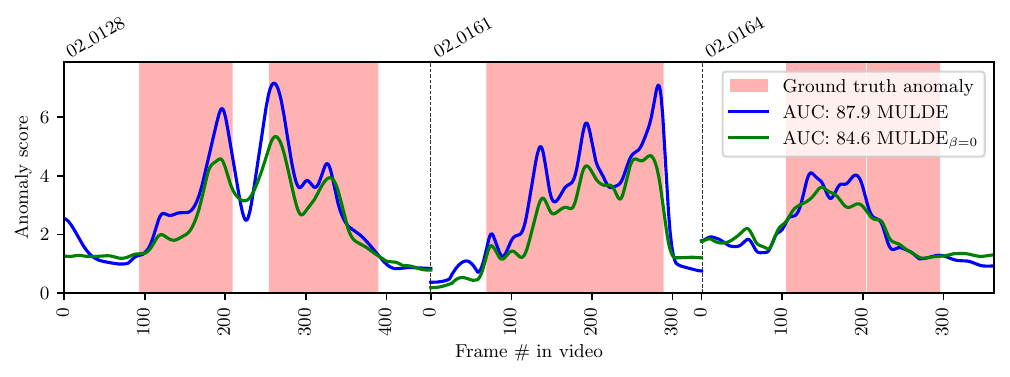}
\caption{The value of MULDE's anomaly score computed for each frame of the test scene 02 of the \ST{} data set (videos 02\_0128, 02\_0161, and 02\_0164) and the resulting \emph{micro} AUC score. 
MULDE with regularization has an advantage (+3.3\%) over the non-regularized MULDE$_{\beta = 0}$.
}
\label{fig:sup:anomaly_scores_scene_st_02}
\end{figure*}

\begin{figure*}
\centering
\includegraphics[width=1.\linewidth]{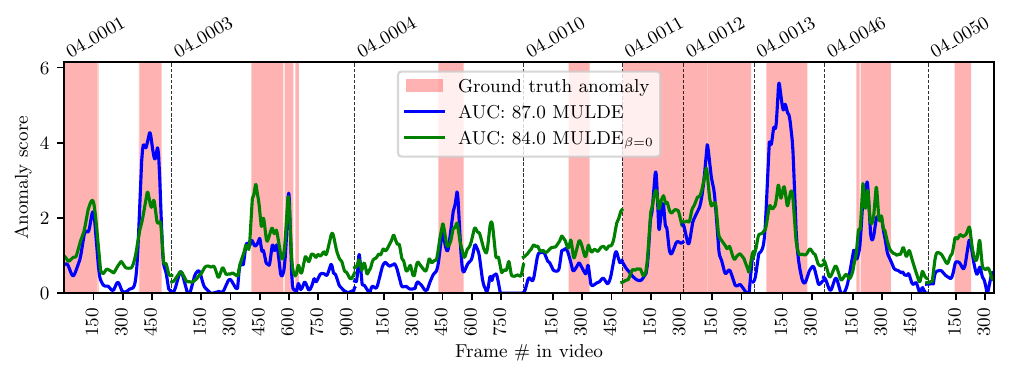}
\caption{MULDE's anomaly score computed for each frame of the test scene 04 of the \ST{} data set (videos 04\_0001, 04\_0003, etc.) and the resulting \emph{micro} AUC score.
MULDE with regularization outperforms the non-regularized MULDE by 3 percent points.
}
\label{fig:sup:anomaly_scores_scene_st_04}
\end{figure*}

\paragraph{Consistency of the anomaly score across different videos}
In Figures~\ref{fig:sup:anomaly_scores_scene_st_02} and~\ref{fig:sup:anomaly_scores_scene_st_04}, we illustrate the trajectories of MULDE's anomaly score across different videos of the same scene, taken from the \ST{} data set.
The levels of the anomaly score for normal fragments of different videos are consistent, as are the levels of the score for anomalous fragments.
The regularized MULDE exhibits better discrimination between normal and anomalous behavior than MULDE$_{\beta=0}$ without regularization. 

\section{Additional Results} \label{sec:ablations}
In this section, we complement the results presented in the main manuscript for the frame-centric and object-centric setup. Furthermore, we provide further details on the choice of $\sigma$, the selection of $L$, and alternatives to the GMM fitting.
\subsection{Frame-centric}
\begin{table}
\centering
\footnotesize
\setlength{\tabcolsep}{5pt}
\begin{tabular}{l *{2}{p{.5cm}}  *{2}{p{.5cm}} *{2}{p{.6cm}}}

\toprule
\multirow{2}{*}{Method}  & \multicolumn{2}{c}{ShanghaiTech}  &   \multicolumn{2}{c}{UCF-Crime}  &   \multicolumn{2}{c}{UBnormal}  \\
\cmidrule(r){2-3} \cmidrule(r){4-5} \cmidrule(r){6-7}
 & Micro & Macro & Micro & Macro & Micro & Macro \\
\midrule
               MNAD-Recon. \cite{park2020learning} &               70.5 &                  - &                - &                - &                - &                - \\
                 Mem-AE. \cite{gong2019memorizing} &               71.2 &                  - &                - &                - &                - &                - \\
                  Frame-Pred. \cite{liu2018future} &               72.8 &                  - &                - &                - &                - &                - \\
              ClusterAE \cite{chang2020clustering} &               73.3 &                  - &                - &                - &                - &                - \\
                    AMMCN \cite{cai2021appearance} &               73.7 &                  - &                - &                - &                - &                - \\
                         MPN \cite{lv2021learning} &               73.8 &                  - &                - &                - &                - &                - \\
                    DLAN-AC \cite{yang2022dynamic} &               74.7 &                  - &                - &                - &                - &                - \\
                           BMAN \cite{lee2019bman} &               76.2 &                  - &                - &                - &                - &                - \\
             CT-D2GAN \cite{feng2021convolutional} &               77.7 &                  - &                - &                - &                - &                - \\
                       CAC~\cite{wang2020cluster} &   \underline{79.3} &                  - &                - &                - &                - &                - \\
                  Scene-Aware~\cite{sun2020scene} &               74.7 &                  - &             72.7 &                - &                - &                - \\
                         BODS~\cite{wang2019gods} &                  - &                  - &             68.3 &                - &                - &                - \\
                         GODS~\cite{wang2019gods} &                  - &                  - &             70.5 &                - &                - &                - \\
                  GCL~\cite{zaheer2022generative} &                  - &                  - &             74.2 &                - &                - &                - \\
     UBnormal~\cite{ubnormal_Acsintoae_CVPR_2022} &                  - &                  - &                - &                - &             68.5 &             80.3 \\
                        FPDM~\cite{Yan_2023_ICCV} &               78.6 &                  - &             74.7 &                - &             62.7 &                - \\
AccI-VAD$_{\text{GMM}}$*~\cite{reiss2022attribute} &               76.2 &               82.9 &             60.3 &             84.5 &             66.8 &             83.2 \\
AccI-VAD$_{\text{kNN}}$*~\cite{reiss2022attribute} &               71.9 &               83.1 &             53.0 &             82.7 &             65.2 &             82.5 \\
               MSMA*~\cite{mahmood2020multiscale} &               76.7 &               84.2 &             64.5 &             83.4 &             70.3 &             85.1 \\
               \hline
                          MULDE$_{\beta=0}$(ours) &               78.4 &      \textbf{86.0} & \underline{75.9} & \underline{84.8} & \underline{71.3} &    \textbf{86.0} \\
                                      MULDE(ours) &      \textbf{81.3} &   \underline{85.9} &    \textbf{78.5} &    \textbf{84.9} &    \textbf{72.8} & \underline{85.5} \\
\bottomrule                            

\end{tabular}
\caption{Frame-centric results. Frame-level AUC-ROC (\%) comparison (best marked \textbf{bold}, second best \underline{underlined}). *implemented by us.}
\label{tab:sup:results_frame_centric}
\vspace{0.6cm}
\end{table}

In Table~\ref{tab:sup:results_frame_centric}, we extend Table~\ref{tab:results_frame_centric} of the main manuscript to include the less recent frame-centric VAD methods.
The results of the competing methods were reproduced after the original publications, except for MSMA~\cite{mahmood2020multiscale} which we re-implemented for processing videos, and AccI-VAD~\cite{reiss2022attribute}, which we adapted to frame-centric operation.
Our method, MSMA, and AccI-VAD used the Hiera-L~\cite{hiera_ryali2023hiera} features.
MULDE surpasses the baselines on all three data sets in terms of both the \emph{micro} and the \emph{macro} metric.

\paragraph{The choice of feature extractors}

\begin{table*}
\centering
\footnotesize
\setlength{\tabcolsep}{5pt}
\begin{tabular}{llcccccccccccccc}
\toprule
\multirow{3}{*}{Dataset} & \multirow{3}{*}{Features} & \multicolumn{7}{c}{Micro} & \multicolumn{7}{c}{Macro} \\
\cmidrule(r){3-9} \cmidrule(r){10-16}
& & \multicolumn{4}{c}{MULDE} & \multirow{2}{*}{MSMA*~} &  \multicolumn{2}{c}{AccI-VAD*} & \multicolumn{4}{c}{MULDE} & \multirow{2}{*}{MSMA*} &  \multicolumn{2}{c}{AccI-VAD*} \\
\cmidrule(r){3-6} \cmidrule(r){10-13}
& & $\beta_{0}$ & $\beta_{0.01}$ & $\beta_{0.1}$ & $\beta_{1}$  & & GMM & kNN & $\beta_{0}$ & $\beta_{0.01}$ & $\beta_{0.1}$ & $\beta_{1}$ & & GMM & kNN \\
\cmidrule(r){1-2} \cmidrule(r){3-9} \cmidrule(r){10-16}
ShanghaiTech &    Hiera-B &          72.1 & \underline{73.4} &             73.1 &    \textbf{73.7} &           71.5 &          67.9 &          68.2 & \underline{83.1} &    \textbf{83.6} &             82.7 &          81.9 &             79.8 &          80.6 &             79.0 \\
ShanghaiTech &    Hiera-L &          78.4 &             80.7 & \underline{81.3} &    \textbf{81.4} &           76.7 &          76.2 &          71.9 &    \textbf{86.0} &             84.9 & \underline{85.9} &          84.5 &             84.2 &          82.9 &             83.1 \\
ShanghaiTech &    Hiera-H &          79.4 &             79.8 & \underline{79.9} &    \textbf{81.7} &           77.4 &          74.7 &          72.7 &    \textbf{88.3} &             87.0 & \underline{87.6} &          86.8 &             86.4 &          86.0 &             83.5 \\
\cmidrule(r){1-2} \cmidrule(r){3-9} \cmidrule(r){10-16}
    UBnormal &    Hiera-B &          70.2 & \underline{71.8} &             71.6 &    \textbf{72.4} &           70.7 &          66.0 &          63.1 &             84.0 & \underline{84.1} & \underline{84.1} &          83.7 &    \textbf{85.4} &          83.6 &             81.9 \\
    UBnormal &    Hiera-L &          71.3 &    \textbf{72.9} & \underline{72.8} &             72.5 &           70.3 &          66.8 &          65.2 &    \textbf{86.0} &             85.2 & \underline{85.5} &          84.7 &             85.1 &          83.2 &             82.5 \\
    UBnormal &    Hiera-H &          70.5 & \underline{72.7} & \underline{72.7} &    \textbf{72.8} &           71.2 &          67.7 &          63.0 & \underline{86.9} &    \textbf{87.1} &    \textbf{87.1} &          86.3 & \underline{86.9} &          85.7 &             84.5 \\
\cmidrule(r){1-2} \cmidrule(r){3-9} \cmidrule(r){10-16}
   UCF-Crime &    Hiera-B & \textbf{74.2} &             72.2 &             71.9 & \underline{72.4} &           69.2 &          69.4 &          68.1 &             85.1 &    \textbf{85.6} & \underline{85.2} &          84.6 &             83.5 &          85.1 &             84.0 \\
   UCF-Crime &    Hiera-L &          75.9 &             76.6 &    \textbf{78.5} & \underline{77.2} &           64.5 &          60.3 &          53.0 &             84.8 & \underline{84.9} & \underline{84.9} & \textbf{85.5} &             83.4 &          84.5 &             82.7 \\
   UCF-Crime &    Hiera-H &          74.8 &    \textbf{76.7} & \underline{75.0} &             74.9 &           71.4 &          60.4 &          57.3 &    \textbf{87.4} &             85.4 &             85.0 &          85.0 & \underline{86.7} &          84.6 &             82.7 \\
   UCF-Crime &        I3D &          67.6 & \underline{70.8} &             69.9 &    \textbf{71.3} &           68.2 &          63.5 &          64.2 &    \textbf{87.6} & \underline{87.5} &             87.3 &          86.5 &             87.1 & \textbf{87.6} & \underline{87.5} \\
\bottomrule
\end{tabular}
\caption{
Frame-level AUC-ROC (\%) comparison. For each input feature representation Hiera-B(ase), Hiera-L(arge), Hiera-H(uge), and I3D, we mark the best scores \textbf{bold} and \underline{underline} the second-best. *adapted from image-based anomaly detection (MSMA) and object-centric VAD (AccI-VAD) to frame-centric VAD.}
\label{tab:sup:ablation_frame_centric}
\end{table*}

In Table~\ref{tab:sup:ablation_frame_centric}, we compare the performance of MULDE used with different video feature extractors in frame-centric VAD.
For reference, the results reported in the main manuscript were obtained with Hiera-L~\cite{hiera_ryali2023hiera}.
We observe, that Hiera-H outperforms Hiera-L in certain experiments, but runs considerably slower.
Hiera-B is the fastest feature extractor, but produces less discriminative features than Hiera-L and Hiera-H. 
Consequently, we opted for Hiera-L due to its favorable tradeoff between computation time and accuracy.

Comparing the \emph{micro} performance attained by MULDE with I3D features (71.3\%, bottom row of Table~\ref{tab:sup:ablation_frame_centric}) to the results of BODS (68.3\%) and GODS (70.5\%, Table~\ref{tab:sup:results_frame_centric}), which both also use I3D features, we see that MULDE is a favorable anomaly detector. 
In particular, this shows that our approach is truly feature-agnostic and works well with any input feature representation.

\subsection{Object-centric}\label{sec:sup:oc}

\begin{table*}
\centering
\footnotesize
\setlength{\tabcolsep}{5pt}
\begin{tabular}{@{}ccccccccccc@{}}
\toprule

\multirow{3}{*}{P} & \multirow{3}{*}{D} & \multirow{3}{*}{V} & \multicolumn{4}{c}{Micro}  &   \multicolumn{4}{c}{Macro} \\ 
\cmidrule(r){4-7} \cmidrule(r){8-11}
& & &  AccI-VAD &  \multirow{2}{*}{MSMA$^*$} & \multirow{2}{*}{MULDE} & \multirow{2}{*}{MULDE$_{\beta=0}$} &  AccI-VAD &  \multirow{2}{*}{MSMA$^*$} & \multirow{2}{*}{MULDE} & \multirow{2}{*}{MULDE$_{\beta=0}$}  \\

& & & kNN$_{\text{P, D}}$ GMM$_{\text{V}}$ & & & & kNN$_{\text{P, D}}$ GMM$_{\text{V}}$ & & & \\
\cmidrule(r){1-3} \cmidrule(r){4-7} \cmidrule(r){8-11}
\checkmark   &             &             &             73.8  &               84.2 & \textbf{84.6}    &  \underline{84.5} & 76.2             & \textbf{87.0}    & \underline{86.5} & 86.0              \\
             & \checkmark  &             &             85.4  &   \underline{87.7} & \textbf{89.0}    &             87.5  & 87.7             & 87.9             & \underline{88.0} & \textbf{88.3}     \\
             &             & \checkmark  &             86.0  &               83.6 & \underline{86.6} &     \textbf{87.4} & 89.6             & 87.2             & \underline{91.8} & \textbf{92.7}     \\
\cmidrule(r){1-3} \cmidrule(r){4-7} \cmidrule(r){8-11}
\checkmark   & \checkmark  &             &             89.3  &               89.5 & \textbf{91.5}    &  \underline{90.6} & 88.8             & \underline{89.5} & \textbf{91.0}    & \textbf{91.0}     \\
             & \checkmark  & \checkmark  &  \underline{93.0} &               89.4 & \textbf{93.1}    &             92.5  & \textbf{95.5}    & 91.2             & \underline{94.7} & \underline{94.7}  \\
\checkmark   &             & \checkmark  &             87.8  &               86.7 & \textbf{91.1}    &  \underline{90.2} & 93.0             & 90.5             & \underline{95.3} & \textbf{95.8}     \\
\cmidrule(r){1-3} \cmidrule(r){4-7} \cmidrule(r){8-11}
\checkmark   & \checkmark  & \checkmark  &  \underline{93.3} &               90.2 & \textbf{94.3}    &             93.1  & \textbf{96.2}    & 92.5             & \underline{96.1} & \underline{96.1}  \\
\cmidrule(r){1-3} \cmidrule(r){4-7} \cmidrule(r){8-11}
\multicolumn{3}{c}{Best}                &  \underline{93.3} &                            90.2 & \textbf{94.3}    &             93.1  & \textbf{96.2}    & 92.5             & \underline{96.1} & \underline{96.1}  \\
\bottomrule
\end{tabular}
\caption{Detailed results for object-centric setup for the \textbf{Avenue} dataset on the \textbf{micro} and \textbf{macro} frame-level AUC-ROC evaluation. Combinations of the object-centric pose (P), deep features (D), and velocities (V) features following the AccI-VAD~\cite{reiss2022attribute} ablations. For every object-centric feature and its combinations, we mark the best scores \textbf{bold} and \underline{underline} the second-best. *adapted from image-based anomaly detection (MSMA) to object-centric VAD.
}
\label{tab:sup:ablations_avenue}
\end{table*}

\begin{table*}
\centering
\footnotesize
\setlength{\tabcolsep}{5pt}
\begin{tabular}{@{}ccccccccccc@{}}
\toprule

\multirow{3}{*}{P} & \multirow{3}{*}{D} & \multirow{3}{*}{V} & \multicolumn{4}{c}{Micro}  &   \multicolumn{4}{c}{Macro} \\ 
\cmidrule(r){4-7} \cmidrule(r){8-11}
& & &  AccI-VAD &  \multirow{2}{*}{MSMA$^*$} & \multirow{2}{*}{MULDE} & \multirow{2}{*}{MULDE$_{\beta=0}$} &  AccI-VAD &  \multirow{2}{*}{MSMA$^*$} & \multirow{2}{*}{MULDE} & \multirow{2}{*}{MULDE$_{\beta=0}$}  \\

& & & kNN$_{\text{P, D}}$ GMM$_{\text{V}}$ & & & & kNN$_{\text{P, D}}$ GMM$_{\text{V}}$ & & & \\
\cmidrule(r){1-3} \cmidrule(r){4-7} \cmidrule(r){8-11}
\checkmark   &             &             & 74.5             & \underline{76.4} & \textbf{78.5}    & 76.0             & 81.0             & \underline{82.2} & \textbf{83.6}    & 82.1             \\
             & \checkmark  &             & 72.5             & 74.6             & \textbf{76.6}    & \underline{74.9} & \underline{82.5} & 78.8             &            82.3  & \textbf{83.0}    \\
             &             & \checkmark  & \textbf{84.4}    & 81.5             &            82.0  & \underline{82.4} & 84.8             & 86.1             & \underline{88.1} & \textbf{88.2}    \\
\cmidrule(r){1-3} \cmidrule(r){4-7} \cmidrule(r){8-11}
\checkmark   & \checkmark  &             & 76.7             & \underline{81.5} & \textbf{82.6}    & 80.5             & 84.9             & \textbf{89.5}    & \underline{88.8} & 87.5             \\
             & \checkmark  & \checkmark  & \textbf{84.5}    & 79.3             & \underline{82.2} & 81.7             & \textbf{88.7}    & 83.5             & \underline{87.9} & 87.4             \\
\checkmark   &             & \checkmark  & 85.9             & 84.1             & \textbf{86.6}    & \underline{86.4} & 88.8             & 90.0             & \textbf{91.5}    & \underline{91.0} \\
\cmidrule(r){1-3} \cmidrule(r){4-7} \cmidrule(r){8-11}
\checkmark   & \checkmark  & \checkmark  & \underline{85.1} & 83.7             & \textbf{86.7}    & 84.8             & 89.6             & \underline{90.2} & \textbf{90.6}    & 89.8             \\
\cmidrule(r){1-3} \cmidrule(r){4-7} \cmidrule(r){8-11}
\multicolumn{3}{c}{Best}                 & 85.9             & 84.1             & \textbf{86.7}    & \underline{86.4} & 89.6             & 90.2             & \textbf{91.5}    & \underline{91.0} \\
\bottomrule
\end{tabular}
\caption{Detailed results for object-centric setup for the \textbf{ShanghaiTech} dataset on the \textbf{micro} and \textbf{macro} frame-level AUC-ROC evaluation. Combinations of the object-centric pose (P), deep features (D), and velocities (V) features following the AccI-VAD~\cite{reiss2022attribute} ablations. For every object-centric feature and its combinations, we mark the best scores \textbf{bold} and \underline{underline} the second-best. *adapted from image-based anomaly detection (MSMA) to object-centric VAD.}
\label{tab:sup:ablations_shanghaitech}
\end{table*}

In the object-centric experiments, reported in Table~\ref{tab:results_object_centric} of the main manuscript, we used MULDE in combination with the feature extraction pipeline proposed by~\citet{reiss2022attribute}. It detects objects in each video frame and extracts deep, velocity, and human pose features for each detected object, as detailed in the manuscript.

Here, we complement these results with the performance attained by MULDE, MSMA~\cite{mahmood2020multiscale}, and AccI-VAD~\cite{reiss2022attribute} using the pose (P), deep (D), and velocity (V) features separately.
AccI-VAD pools the highest anomaly scores from each frame for P, D, and V and normalizes each feature type by its min/max training counterpart. Finally, the scores of P, D, and V are added up using pairs of the feature types and the triplet. MULDE differs in that regard, instead of min-/max-normalization, we standardize by the training statistics, then clip negative values which are normal, and add up. 

We follow AccI-VAD's ablation protocol for MULDE and present the results for the \Avenue{} data set in Table~\ref{tab:sup:ablations_avenue}.
Table~\ref{tab:sup:ablations_shanghaitech} contains the results obtained for \ST{}. 
For \Avenue{}, the combination of all three feature types gives the best micro and macro scores.
Similarly, for \ST{}, P, D, V leads to the best micro score, and the combination of P and V leads to the best macro score.
These results show that it might be beneficial to combine features, encoding complementary information. MULDE makes this combination easy as it is feature-agnostic. 

\paragraph{Performance in terms of the region- and track-based detection criteria}
\begin{table}
\centering
\footnotesize
\setlength{\tabcolsep}{5pt}
\begin{tabular}{lllll}
\toprule
\multirow{2}{*}{Method} & \multicolumn{2}{c}{Avenue} & \multicolumn{2}{c}{ShanghaiTech} \\
\cmidrule(r){2-3} \cmidrule(r){4-5}
& RBDC & TBDC & RBDC & TBDC \\
\midrule
Ramachandra \etal \cite{ramachandra2020street}                                       &              35.8 &     \textbf{80.9} &                 - &                 - \\
Ramachandra \etal \cite{ramachandra2020learning}                                       &              41.2 &              78.6 &                 - &                 - \\
 Frame-Pred. \cite{liu2018future}                                                      &                 - &                 - &              17.0 &              54.2 \\
 CAE-SVM \cite{ionescu2019object}                                                      &                 - &                 - &              20.6 &              44.5 \\
 BA-AED \cite{georgescu2021background}                                                 &              65.0 &              67.0 &              41.3 &              78.8 \\
 SSMTL \cite{georgescu2021anomaly}                                                     &              57.0 &              58.3 &              42.8 &              83.9 \\
 SSMTL \cite{georgescu2021anomaly}+UBnormal \cite{ubnormal_Acsintoae_CVPR_2022}$^\dagger$        &              61.1 &              61.4 &              47.2 &     \textbf{86.2} \\
 \midrule
 MULDE$_{\text{D}}$ (ours)                                                            &     \textbf{73.1} &              74.4 &              48.9 &              81.2 \\
 MULDE$_{\text{V}}$ (ours)                                                            &              13.8 &              46.8 &     \textbf{55.0} &  \underline{85.6} \\
 MULDE$_{\text{D, V}}$ (ours)                                                         &  \underline{71.8} &  \underline{79.2} &  \underline{52.7} &              83.6 \\
\bottomrule
\end{tabular}
\caption{Localization-based evaluation using RBDC and TBDC scores \cite{ramachandra2020street}. We provide scores for regions based on deep features (D) only, velocity (V) only and the combination of D, V.
$^\dagger$extended training data used.
}
\label{tab:sup:rbdc_tbdc}
\end{table}

The region- and track-based detection criteria (RBDC and TBDC) were introduced by \citet{ramachandra2020street} to assess the anomaly localization capabilities of VAD methods, which are not captured by the more common frame-level AUC-ROC scores. 

RBDC and TBDC require pixel-level anomaly scores. AccI-VAD~\cite{reiss2022attribute}, SSMTL~\cite{georgescu2021anomaly}, BA-AED~\cite{georgescu2021background}, MULDE apply the anomaly score to the bounding-box, \ie the region obtained by the object detector. We computed the RBDC and TBDC metrics for MULDE using the code and annotations released by~\citet{georgescu2021background}.
We provide RBDC and TBDC scores for regions based on deep features (D) only, velocity (V) only, and the combination of D and V in Table~\ref{tab:sup:rbdc_tbdc}.
Pose (P) is not used for this evaluation, as the features provided by~\citet{reiss2022attribute} are already normalized to the top left image corner and thus, can not be attributed to a specific location within the frame.

For comparison, we report the results of the baseline methods after~\cite{ubnormal_Acsintoae_CVPR_2022}.
MULDE clearly outperforms previous approaches in terms of the region-based RBDC. 
In terms of the TBDC, MULDE is among the top-performing approaches, outperformed only by~\cite{ramachandra2020street} on \Avenue{} and by~\cite{ubnormal_Acsintoae_CVPR_2022} (which requires additional training data) on \ST{}.

\subsection{Parameter selection and alternatives to GMM}
In this section, we discuss the selection of the noise range $\sigma$, the selection of the number of 
noise scales $L$, details on the GMM fitting and alternative approaches to the GMM fitting.

\paragraph{Noise range selection of $\sigma$}
Even though our method eliminates the need to select the noise range $[\sigma_{\text{low}}, \sigma_{\text{high}}]$ used for training, the range of employed noise scales should be sufficiently large to cover anomalies that would be seen at test time. Currently, there is no automatic way to select the upper limit of the noise range. 
We circumvent this limitation by standardizing the features component-wise and using a fixed, wide range of noise scales. We set $\sigma_\text{high}=1.0$ to make it equal to the standard deviation of the distribution of training video features. The $\sigma_\text{low}=0.001$ was selected to make the interval wide. We kept this range for all data sets, even though Figure~\ref{fig:sigma_sweep_gmm} of the main manuscript (frame-centric micro score on \ST{}) suggests that such a wide interval might not be necessary: When used with a single $\sigma$, MULDE performs best for $\sigma=0.33$, and $\sigma<0.2$ or $\sigma>0.5$ lead to much lower scores. Initial experiments showed promising results across all the datasets; thus, we did not fine-tune these hyperparameters.

\paragraph{Selection of number of noise scales $L$}
In all the reported experiments, we decimated the range of noise scales into $L=16$ points.
Testing other values of $L$ in the frame-centric VAD on \Ubnormal{} (results reported in Table~\ref{tab:sup:L}) reveals that MULDE is not sensitive to the number of noise scales used. 

\begin{table}
\centering
\footnotesize
\setlength{\tabcolsep}{4pt}
\begin{tabular}{r | c c c c c}
\toprule
$L$          &    4 &    8 &    16 &    32 &    64 \\
\midrule
AUC-ROC & 72.16& 72.80& 72.89 & 72.95 & 72.99 \\
\bottomrule
\end{tabular}
\caption{Frame-centric results on \Ubnormal{} for a different number of noise scales $L$. Frame-level micro AUC-ROC (\%).}
\label{tab:sup:L}
\end{table}

\paragraph{Details on GMM fitting}
Once the network $f_{\theta}$ is trained, we compute the multi-scale log-density approximation for each video feature $\mathbf{x}$ in the training set $\mathcal{T}$. This results in a data set of vectors $\{[ f_\theta (\mathbf{x},\sigma_1), \ldots f_\theta(\mathbf{x},\sigma_L) ]\}_{\mathbf{x}\in\mathcal{T}}$. In other words, each single $d-$dimensional video feature $\mathbf{x}$ is evaluated at $L$ noise scales which results in a new $L-$dimensional feature vector. We then fit a $L-$dimensional GMM with one, three, and five components to this set of vectors using the Expectation–maximization algorithm. 

At test time, our neural network takes a vector of a video feature and produces a multi-scale vector of log-density approximations, which is then input to the GMM yielding a negative log-likelihood which we use as the anomaly score. Finally, like in previous work~\cite{ubnormal_Acsintoae_CVPR_2022, reiss2022attribute, BARBALAU2023103656, ionescu2019object, georgescu2021background, georgescu2021anomaly}, these scores are temporally smoothed with a 1d-Gaussian filter to obtain the final anomaly score.

\paragraph{Alternatives to GMM}
As discussed in section~\ref{sec:method:multiscale} of the manuscript, in theory, a log-likelihood estimation at a well-chosen noise level is sufficient to detect anomalies. 
This is confirmed by the result presented in Figure~\ref{fig:sigma_sweep_gmm} of the manuscript, where we see that using the best, single noise level yields a micro score on par with the GMM.
However, the choice of the optimal noise scale is not trivial: the noise should be high enough to blend modes of the probability density function originating from individual training samples but not so high as to distort the shape of the original, noise-free distribution.
It is difficult to determine the optimal noise level without anomalous validation data, which prompted us to use the GMM.  
Theoretically, we could substitute the GMM with an alternative aggregation method, for example,  max-, average-, or median-pooling.
We evaluated these methods as follows:
Before pooling, we equalized the log-density estimates by standardizing them across each noise scale with their respective means and standard deviations computed over the training set.
We evaluate the alternative pooling method to the frame-centric experiment on \ST{}. As reported in Table~\ref{tab:results_frame_centric} in the manuscript, MULDE$_{\beta=0}$ with the GMM attains a Micro score of 78.4.
This result decreases to 76.30 with max-pooling, 76.02 with average-pooling, and 75.83 with median-pooling instead of the GMM.

\end{document}